\documentclass{article}

     \usepackage[preprint]{neurips_2019}

\usepackage[utf8]{inputenc} 
\usepackage[T1]{fontenc}    
\usepackage{hyperref}       
\usepackage{url}            
\usepackage{booktabs}       
\usepackage{amsfonts}       
\usepackage{nicefrac}       
\usepackage{microtype}      

\usepackage{amsthm,amsmath,amssymb}
\usepackage{algorithm, algorithmic}
\usepackage{cleveref}
\usepackage{subfiles}

\usepackage{mathtools}
\usepackage{commath}

\usepackage{silence}
\let\norm\undefined 
\DeclarePairedDelimiter\norm{\lVert}{\rVert}
\DeclareMathOperator*{\argmin}{arg\,min}

\newcommand{\Z}{\mathcal{Z}}

\newcommand{\X}{\mathcal{X}}

\newcommand{\G}{\mathcal{G}}

\usepackage{color}

\newcommand{\K}{\mathcal{K}}

\newcommand{\E}{\mathbb{E}}

\newcommand{\reals}{\mathbb{R}}

\newtheorem{definition}{Definition}[section]
\newtheorem{theorem}[definition]{Theorem}

\newtheorem{lemma}[definition]{Lemma}
\newtheorem{corollary}[definition]{Corollary}

\newcommand{\smaller}[1]{{\mathchoice{}{}{\scriptscriptstyle}{}#1}}

\newcommand{\linspace}{\mathcal{Z}}
\newcommand{\lininputspace}{\mathcal{X}}
\newcommand{\linoutputspace}{\mathcal{Y}}
\newcommand{\point}{z}
\newcommand{\inputpoint}{x}
\newcommand{\outputpoint}{y}
\newcommand{\fclass}{\mathcal{G}}
\newcommand{\f}{g}
\newcommand{\probdistrib}[1]{\mathbb{#1}}
\newcommand{\empprobdistrib}[1]{\hat{\mathbb{#1}}}
\newcommand{\ipm}[3]{\gamma_{#3}(#1,#2)}
\newcommand{\ipmdef}[2]{\sup_{\f\in\fclass}\left|\E_{#1} \f(\point) -\E_{#2} \f(\point)\right|}
\newcommand{\genindex}{i}
\newcommand{\numsources}{J}
\newcommand{\sourceindex}{j}
\newcommand{\targetindex}{\smaller{T}}
\newcommand{\sample}{Z}

\newcommand{\sourcesamplegen}{\sample^{(\sourceindex)}}

\newcommand{\sourcesamplepointgen}{\point^{(\sourceindex)}_\genindex}

\newcommand{\targetsample}{\sample^{\targetindex}}
\newcommand{\targetsamplepoint}{\point^{\targetindex}_\genindex}
\newcommand{\sourcedistrib}{\probdistrib{S}^{(\sourceindex)}}

\newcommand{\targetdistrib}{\probdistrib{T}}
\newcommand{\sourceempdistrib}{\empprobdistrib{S}^{(\sourceindex)}}
\newcommand{\targetempdistrib}{\empprobdistrib{T}}
\newcommand{\sourcemixtureempdistrib}{\empprobdistrib{S}_\alpha}
\newcommand{\model}{f}
\newcommand{\modelclass}{\mathcal{F}}
\newcommand{\loss}{\ell}
\newcommand{\size}[1]{N^{(#1)}}
\newcommand{\modelbasis}{\psi}
\newcommand{\lincoeff}{w}
\newcommand{\lincoeffspace}{\mathcal{W}}
\newcommand{\modelbasisparam}{\theta}
\newcommand{\modelbasisparamspace}{\Theta}
\newcommand{\kernelbasis}{\phi}

\title{Weighted Meta-Learning}

\author{%
  Diana Cai \\
  Princeton University \\
  Princeton, NJ 08544 \\
  \texttt{dcai@cs.princeton.edu} \\
   \And
   Rishit Sheth \\
   Microsoft Research New England \\
   Cambridge, MA 02142 \\
   \texttt{rishet@microsoft.com} \\
   \AND
   Lester Mackey \\
   Microsoft Research New England \\
   Cambridge, MA 02142 \\
   \texttt{lmackey@microsoft.com} \\
   \And
   Nicolo Fusi \\
   Microsoft Research New England \\
   Cambridge, MA 02142 \\
   \texttt{fusi@microsoft.com} \\
}

\begin{document}

\maketitle

\begin{abstract}
    Meta-learning leverages related source tasks to learn an initialization that
    can be quickly fine-tuned to a target task with limited labeled examples.
    However, many popular meta-learning algorithms, such as model-agnostic
    meta-learning (MAML), only assume access to the target samples for
    fine-tuning. In this work, we provide a general framework for meta-learning
    based on weighting the loss of different source tasks, where the weights are
    allowed to depend on the target samples. In this general setting, we provide
    upper bounds on the distance of the weighted empirical risk of the source
    tasks and expected target risk in terms of an integral probability metric
    (IPM) and Rademacher complexity, which apply to a number of meta-learning
    settings including MAML and a weighted MAML variant.
    We then develop a learning algorithm based on minimizing the error bound
    with respect to an empirical IPM, including a weighted MAML algorithm, $\alpha$-MAML.
    Finally, we demonstrate empirically on several regression
    problems that our weighted meta-learning algorithm is able to find better
    initializations than uniformly-weighted meta-learning algorithms, such as
    MAML.
\end{abstract}


\section{Introduction}

The applicability of machine learning techniques to real-world problems is often
limited by the quantity of labeled data available. This is particularly
detrimental when high-accuracy, high-capacity models are needed for a given application, since their
requirements on the amount of data are particularly onerous. As a result,
examples of these issues
are wide-ranging and can be identified in vision \citep{koch2015siamese}, language modeling
\citep{vinyals2016matching}, content recommendation \citep{vartak2017meta}, character generation
\citep{lake2015human}, and health care \citep{zhang2019metapred, altae2017low}.

One crucial observation to overcome this challenge is that while data on the \emph{target} task may
be limited, other \emph{source} tasks can be used to help with learning. This is precisely the
setting considered in meta-learning, wherein multiple source tasks are used to provide a good
``initialization'' to learn on a target task.
Some recent developments in meta-learning include
metric-based methods \citep{lake2015human,koch2015siamese,vinyals2016matching,snell2017prototypical,oreshkin2018tadam},
model-based methods \citep{santoro2016meta,munkhdalai2017meta},
optimization-based methods \citep{ravi2016optimization},
and gradient-based methods \citep{finn2017model,nichol2018first}.

In gradient-based meta-learning, the goal is to learn an initialization from a
set of source tasks that can be quickly adapted to a new target task with a small number of gradient steps. Within gradient-based meta-learning methods, model-agnostic meta-learning (MAML) \citep{finn2017model} is a popular
approach that leverages data from a collection of source tasks to learn an initial model that can be
quickly adapted to some target data task, often using a limited number of
labeled target examples.
A key feature of MAML is that it does not require a particular type of learning model
or architecture and is therefore broadly applicable to problems in regression, classification,
and reinforcement learning. A number of extensions to MAML
\citep{nichol2018first,antoniou2018train,song2020maml} have since been proposed,
and connections to hierarchical Bayesian modeling have been drawn \citep{grant2018recasting,yoon2018bayesian,finn2018probabilistic,ravi2019amortized,jerfel2019reconciling}.

An important assumption in many gradient-based
meta-learning methods is that the source and target tasks are drawn from the same
task distribution.
Since the true task distribution is usually unknown, implicit in this assumption
is that future target tasks will be
uniformly similar to the source tasks.
In practice, this assumption is encoded in the algorithm
as uniformly sampling from the source tasks during meta-training
\citep{finn2017model,nichol2018first}.
However,
a target task may be similar to only a few of the source tasks, or even just one, and
applying equal weighting to all sources during meta-learning can be detrimental.
Indeed, recent research in extending the MAML framework by modeling hierarchical
task distributions \citep{yao2019hierarchically},
task non-stationarity \citep{nagabandi2018deep},
and multi-modality \citep{vuorio2018toward}
attempts to address this shortcoming with more complex meta-learners,
and other meta-learning methods
have noted the importance task similarity
\citep{achille2019task2vec, jomaa2019dataset2vec}.
Here, we instead note that in many practical applications,
the target task is available during training, and focus
on the goal of minimizing the loss of the specific target task during the
\emph{entire} training procedure, rather than just the adaptation step.
Specifically, we propose using the labeled target task samples
during meta-training to learn a better initialization for a given target task.

We study a general class of meta-learning methods
that can be described by a task-weighted meta-objective;
this general class
captures a variety of gradient-based meta-learning
objectives, such as joint training and MAML (and first-order variants), as well as weighted variants of
joint training and MAML.
We make no assumptions on the distribution of the
source and target tasks.
Our meta-objective is designed to
encode similarity between the task distributions
by upweighting sources that are more similar to the
target task.
The similarity between source and target task distributions
is captured by an integral probability metric (IPM),
which is used to compare the empirical distributions
of the tasks.

For this class of weighted meta-learning objectives,
we provide data-dependent error bounds on the
expected target risk
in terms of an empirical IPM and Rademacher complexity.
The resulting generalization bound leads naturally to a
learning algorithm incorporating weight optimization.
We show that the IPM calculation can be bounded by selecting a kernel that generates
a reproducing kernel Hilbert space (RKHS) ball containing
the class of functions described by composing the model class with the loss function.
We provide examples on how to construct such an RKHS ball
for squared loss (regression) and hinge loss (binary classification) with linear
basis function models,
which apply to weighted MAML and weighted ERM.
Importantly, this approach defines task similarity
explicitly in terms of performance rather than a proxy measure,
e.g.,
task embedding distances.

In what follows, we first review related work that has considered
task similarity in meta-learning
(\Cref{sec-related}).
We then describe our general meta-learning setup
(\Cref{ssec-setting})
and present data-dependent generalization bounds
(\Cref{sec-weighted-maml-bounds}).
An algorithm for minimizing the weights
of the meta-objective that is used to learn the initial model
is described in \Cref{sec-algorithm}.
Finally,
we empirically demonstrate that a weighted
meta-learning objective can lead to improved
intitializations over uniformly-weighted meta-learning
objectives, which include joint training and MAML as
special cases.
In particular,
we conduct experiments in synthetic linear and sine regression problems,
as well as a number of multi-dimensional basis regression problems on real data
sets (\Cref{sec-experiments}).


\section{Related work}
\label{sec-related}

A number of recent works have established guarantees for gradient-based meta-learning algorithms
\citep{finn19a,khodak2019provable,khodak2019adaptive}
developed from the perspective of online convex optimization.
Further work has also established guarantees for non-convex loss functions \citep{fallah2019}.
In these frameworks,
task similarity is either not considered, or is
fundamentally defined as distance between model parameters in some metric space
\citep{khodak2019provable,khodak2019adaptive},
whereas we define task similarity via an IPM that directly captures induced performance differences.
\citet{li2017meta,xu2019} incorporate
the use of task-weighted loss functions within MAML meta-training.
However, the weights are found heuristically with no guarantees and
are not explicitly related to task similarity.

A separate line of work in domain adaptation studies the problem of combining multiple source tasks with target task data.
Early bounds for classification were established by \citet{ben2010theory} in terms of an $\mathcal{H}$-divergence.
\citet{zhang2013arxiv,zhang2012generalization} extend these results by considering
general loss functions and deriving bounds in terms of a population IPM
(and subsequently study convergence in this setting).
Separately, \citet{mansour2009domain} considered the mixture adaptation problem of combining the predictions of given source models and showed that a distribution-weighted combining rule will achieve performance close to the lowest performing source model assuming the target is a mixture of sources.
The ensemble generative adversarial network of \citet{adlam2019learning} utilizes a discrepancy distance \citep{mansour2009bdomain,cortes2014domain} to compute task weights, but utilizes fixed models in the ensemble to generate data for a target task, whereas we learn task weights to optimize a model for a target task directly.
Similar to our setting, \citet{pentinamulti} also develop a data-dependent bound
for meta-learning with weighted tasks; their bound, however, contains interaction terms between task
weights and unobservable quantities (the minimum possible combined source/target error of a single
hypothesis) which, unlike this work, precludes optimization with respect to task weights.

In a similar spirit to our work,
\citet{shui2019principled} consider the $\mathcal{H}$-divergence and Wasserstein distance
as task similarity measures to develop generalization bounds in the setting of
multi-task learning with finite VC- and pseudo-dimension model classes.
In our construction,
we embed the model class composed with loss function within a RKHS,
allowing the task similarity measure to be efficiently computed by kernel distance.

Finally, there are other lines of work that capture
notions of task similarity for meta-learning
through proxy measures such as distance between embedded tasks
\citep{achille2019task2vec,jomaa2019dataset2vec}.

\section{Weighted meta-learning}

\subsection{Setting and objective}
\label{ssec-setting}

Let $\lininputspace$ and $\linoutputspace$ represent input and output spaces
respectively, and define $\linspace :=\lininputspace\times\linoutputspace$.
Suppose we have
$\numsources$ independently drawn \emph{source} tasks
$\{\sourcesamplegen\}_{\sourceindex=1}^\numsources$,
where the $j$-th task
$\sourcesamplegen :=\{\sourcesamplepointgen\}_{\genindex=1}^{\size{\sourceindex}}$
is defined by a set of data points
$\point^{(\sourceindex)}_\genindex\in\linspace$.
We assume the instances $\{\sourcesamplepointgen\}$ of a source $\sourceindex$
are drawn i.i.d.\ from some unknown distribution $\sourcedistrib$
and that the distributions of
the source tasks may be different.

The objective is to use the source tasks
to learn an initial model,
$\model:\lininputspace\rightarrow\linoutputspace$,
that generalizes well with respect to a loss function
$\loss:\linoutputspace\times\linoutputspace\rightarrow\mathbb{R}$
and an unknown \emph{target} distribution $\targetdistrib$ over
$\lininputspace\times\linoutputspace$.
That is, the expected target risk
$\E_{\targetdistrib} \loss(\outputpoint, \model(\inputpoint))$ is small.

Importantly, we assume that a small i.i.d.\ sample from the target distribution,
$\targetsample=\{\targetsamplepoint\}_{\genindex=1}^{\size{\targetindex}}$,
is available and can be utilized during training,
where
$\size{\targetindex} \ll \size{\sourceindex}$,
for all $1\le\sourceindex\le\numsources$.
In the model-agnostic meta-learning (MAML) framework of \citet{finn2017model},
the target sample is utilized during a ``fast adaptation'' phase after learning
an initial model but prior to prediction on the target task.
In constrast, rather than using only the source tasks during meta-training,
we additionally use this labeled target sample $\targetsample$ to learn the initial model.

In the following,
let $\delta_\point$ denote the Dirac measure at $\point\in\linspace$.
Denote the $j$-th empirical source distribution
and
the empirical target distribution by
\begin{align*}
\sourceempdistrib :=
\frac{1}{\size{\sourceindex}}\sum_{\genindex=1}^{\size{\sourceindex}}
    \delta_{\sourcesamplepointgen},
    \qquad
\targetempdistrib :=
\frac{1}{\size{\targetindex}}\sum_{\genindex=1}^{\size{\targetindex}}
    \delta_{\targetsamplepoint},
\end{align*}
respectively.
Given weights $\alpha \in \Delta^{\numsources-1} := \{\alpha \in [0,1]^{\numsources}:
\sum_{\sourceindex=1}^\numsources \alpha_\sourceindex = 1\}$,
we define
the empirical $\alpha$-mixture distribution among the $\numsources$ source
samples as
$\sourcemixtureempdistrib: = \sum_{\sourceindex=1}^\numsources
\alpha_\sourceindex \sourceempdistrib$.

With this notation, the empirical risk of a model on a source task is given by
$\E_{\sourceempdistrib} \loss( \outputpoint, \model(\inputpoint) )$,
the empirical risk on the target task by
$\E_{\targetempdistrib} \loss( \outputpoint, \model(\inputpoint) )$,
and the empirical risk
on an $\alpha$-mixture of source samples by
$\E_{\sourcemixtureempdistrib} \loss( \outputpoint, \model(\inputpoint) )$.

Let $\fclass$ be a function class with members mapping from $\linspace$ to $\mathbb{R}$.
We consider a class of meta-learning algorithms that learns the initial model
by minimizing the following task-weighted meta-objective:
\begin{align}
    \label{eq-meta-objective}
    \sum_{j=1}^J \alpha_j \E_{\sourceempdistrib} g(z),
\end{align}
where $\alpha \in \Delta^{\numsources-1}$.
Let $\loss:\linoutputspace\times\linoutputspace\rightarrow\reals$ denote a loss function
and
$\modelclass=\{\model(\inputpoint;\modelbasisparam): \modelbasisparam\in\modelbasisparamspace\}$
denote a parameterized predictor or model class
with
$\model(\cdot;\modelbasisparam):\lininputspace\rightarrow\linoutputspace$.
Joint training (i.e., standard ERM with uniform weights on the tasks)
is instantiated in this framework with uniform weights
$\alpha_\sourceindex=1/\numsources$
and the function class
\begin{align}
    \label{eq-erm}
\fclass=\left\{\f(\inputpoint,\outputpoint)
= \loss(\outputpoint, \model(\inputpoint; \modelbasisparam)): \modelbasisparam\in\modelbasisparamspace\right\}.
\end{align}
MAML is instantiated with
$\alpha_\sourceindex=1/\numsources$ and
\begin{align}
\label{eq-maml-class}
\fclass=\left\{\f(\inputpoint,\outputpoint)=
    \loss(\outputpoint,\model(\inputpoint; U(\modelbasisparam))):
\modelbasisparam\in\modelbasisparamspace\right\},
\end{align}
where $U$ is an adaptation function defined by
\begin{equation*}
    U(\modelbasisparam):=\modelbasisparam-\eta\nabla_\modelbasisparam\E_{\sourcemixtureempdistrib}\loss(\outputpoint,\model(\inputpoint;\modelbasisparam))
    \label{eq:maml-u-function}
\end{equation*}
and $\eta$ is a global step-size parameter.

In this work,
we will explicitly consider the function classes
in \Cref{eq-erm} and \Cref{eq-maml-class}
with non-uniform  weights $\alpha \in \Delta^{J-1}$,
but other function classes can also be considered in this framework,
including the gradient-based meta-learning methods
first-order MAML and Reptile \citep{nichol2018first}.
However, note that the bound presented in \Cref{thm-second}
applies to general classes of functions $\G$ mapping $\Z$ to $\reals$.

Given the objective of \Cref{eq-meta-objective},
it might be natural to upweight source tasks
that are more similar to the target task.
In particular, we will use
an integral probability metric (IPM) \citep{muller1997integral}
as a measure of distance between
the distributions of the weighted sources and the target.

\begin{definition}
\label{def-ipm}
The integral probability metric (IPM) between two probability distributions $\probdistrib{P}$ and $\probdistrib{Q}$ on $\linspace$
with respect to the class of real-valued functions $\fclass$ is defined as
\begin{align}
    \label{eq-ipm}
    \ipm{\probdistrib{P}}{\probdistrib{Q}}{\fclass} := \ipmdef{\probdistrib{P}}{\probdistrib{Q}}.
\end{align}
\end{definition}

Many popular metrics between probability distributions can be
cast in terms of an IPM with respect to a specific class of functions $\G$,
such as the total variation distance,
the Wasserstein distance, the bounded Lipschitz distance, and the kernel distance
(c.f.\ \citet[Table~1]{sriperumbudur2012empirical}).

The IPM has been applied to function classes involving
a parameterized model $f$ in a number of different contexts,
including domain adaptation \citep{zhang2012generalization}.
The IPM also describes the
discrepancy distance of \citet{mansour2009bdomain} (c.f.
\citet[Section~2]{adlam2019learning}) for comparing
distributions defined on the input space $\X$,
and other IPMs between distributions
on $\X$ have been used for learning in generative adversarial networks \citep{zhang2018discrimination}.
The discrepancy distance of \citet{mansour2009bdomain}
is itself a generalization of the $\mathcal{H}$-divergence of \citet{ben2010theory}
from 0-1 loss to arbitrary losses.
We refer to \citet[Sec.~3.2]{zhang2013arxiv} for additional discussion of these relationships.

In \Cref{sec-algorithm}, we provide a computable
algorithm for finding the $\alpha$ weight values,
based on computing the IPM between source and target samples
with respect to the class of functions consisting of the composition of the loss and model class.

\subsection{Data-dependent bound for weighted meta-learning}
\label{sec-weighted-maml-bounds}

We now provide a data-dependent upper bound on the
distance between the empirical risk of an $\alpha$-mixture of source tasks
and the expected risk of the target task,
where the bound holds uniformly over the class of functions $\G$:
that is, we upper bound the quantity
\begin{align}
    \label{eq:weighted-source-pop-ipm}
    \ipm{\sourcemixtureempdistrib}{\targetdistrib}{\fclass}
    =
    \sup_{\f\in\fclass} \bigg|
        \sum_{\sourceindex=1}^\numsources
        \alpha_\sourceindex
        \E_{\sourceempdistrib}
        \f(\point)
        -
        \E_{\targetdistrib}
        \f(\point)
    \bigg|.
\end{align}
The bound directly yields (i) a generalization bound for target risk in terms of an empirical
IPM between weighted source samples and the target sample and (ii) a computable algorithm for
finding the weights $\alpha$ that minimize the bound.

We first present a definition and corresponding result that will be used for our
bound.
\begin{definition}
   The empirical Rademacher complexity
    of a function class $\G$ with respect to
    a sample
     $\{z_i\}_{i=1}^N$
    drawn
    i.i.d.\ from a distribution $\mathbb{P}$
    is defined as
    \begin{align*}
        \mathcal{R}(\G | z_1,\ldots,z_{N}) :=
        \E \sup_{g \in \G} \frac{1}{N} \left|\sum_{i=1}^N \sigma_i \f(z_i)\right|,
    \end{align*}
    where the expectation is taken w.r.t.\ the i.i.d.\ Rademacher random variables
    $\{\sigma_i\}$.
    The expected Rademacher complexity is
    defined as
    \begin{align*}
        \mathcal{R}(\G) := \E_{\mathbb{P}^{N}}
        \mathcal{R}(\G | z_1,\ldots,z_{N}),
    \end{align*}
        where
    $\mathbb{P}^{N}$
    denotes an $N$-fold product distribution of $\mathbb{P}$.
\end{definition}

The following is a standard uniform deviation bound based on Rademacher
complexity (c.f.\ \citet{bartlett2002rademacher}):
\begin{lemma}[Uniform deviation with empirical Rademacher complexity]
    \label{lemma-symmetrization}
    Let the sample $\{\point_1,\ldots,\point_N\}$ be drawn i.i.d.\ from a distribution $\probdistrib{P}$ over $\linspace$ 
    and
    let $\fclass$ denote a class of functions on $\linspace$ with members mapping from $\linspace$ to $[a,b]$.
    Then for $\epsilon>0$, we have that with probability at least $1-\epsilon$ over the draw of the sample,
    \begin{align}
        \label{eq-rademacher-symmetrization}
        \sup_{\f \in \fclass} \left| \E_{\empprobdistrib{P}} \f(\point) - \E_{\probdistrib{P}} \f(\point) \right|
        \leq
        2 \,
        \mathcal{R}(\fclass|\point_1,\dots,\point_N)
        +
        3\sqrt{\frac{(b-a)^2\log(2/\epsilon)}{2 N}},
    \end{align}
    where
    $\empprobdistrib{P}$ represents the empirical distribution of the sample, 
    and
    $\mathcal{R}(\fclass|\point_1,\dots,\point_N)$
    denotes the empirical Rademacher complexity of the function class $\fclass$ w.r.t.\ the sample.
\end{lemma}

Our main result is the following data-dependent upper bound
on $\ipm{\sourcemixtureempdistrib}{\targetdistrib}{\fclass}$,
which decomposes into a sum of
the IPM between the empirical distribution of the $\alpha$-mixture of sources $\sourcemixtureempdistrib$
and empirical target distribution $\targetempdistrib$ and
{the empirical Rademacher complexity with respect to the target distribution}.
\begin{theorem}
    \label{thm-second}
    Let $\fclass$ denote a class of functions whose members map from $\linspace$ to $[a,b]$,
    and suppose that the source tasks
    $\{ \sourcesamplegen \}_{\sourceindex=1}^\numsources$ and target task $\targetsample$
    are independent, and that the data instances of each are i.i.d.\ within a sample.
    Let $\epsilon > 0$.
    Then with probability at least $1-\epsilon$ over the draws of the source and target samples,
\begin{align}
    \ipm{\sourcemixtureempdistrib}{\targetdistrib}{\fclass}
    \leq
    \ipm{\sourcemixtureempdistrib}{\targetempdistrib}{\fclass}
    +
    {2\mathcal{R}(\fclass|\point_1,\dots,\point_{\size{\targetindex}})}
    +
    3\sqrt{\frac{(b-a)^2\log(2/\epsilon)}{2\size{\targetindex}}}
    ,
\end{align}
    where
    $\mathcal{R}(\fclass|\point_1,\dots,\point_{\size{\targetindex}})$
    denotes the empirical Rademacher complexity of the function class $\fclass$
    w.r.t.\ the target sample.
\end{theorem}
\begin{proof}
    With probability 1 over the draw of target sample,
    we have
    \begin{align*}
        \ipm{\sourcemixtureempdistrib}{\targetdistrib}{\fclass}
        &=
        \sup_{\f\in\fclass}\left|
            \E_{\sourcemixtureempdistrib} \f(\point)
            +
            \E_{\targetempdistrib} \f(\point)
            -
            \E_{\targetempdistrib} \f(\point)
            -
            \E_{\targetdistrib} \f(\point)
        \right|
        \\
        &\leq
        \sup_{\f\in\fclass}
        \left[
            \left|
                \E_{\sourcemixtureempdistrib} \f(\point)
                -
                \E_{\targetempdistrib} \f(\point)
            \right|
            +
            \left|
                \E_{\targetempdistrib} \f(\point)
                -
                \E_{\targetdistrib} \f(\point)
            \right|
        \right]
        \\
        &\leq
        \sup_{\f\in\fclass}
        \left|
            \E_{\sourcemixtureempdistrib} \f(\point)
            -
            \E_{\targetempdistrib} \f(\point)
        \right|
        +
        \sup_{\f\in\fclass}
        \left|
            \E_{\targetempdistrib} \f(\point)
            -
            \E_{\targetdistrib} \f(\point)
        \right|
        \\
        & =
        \ipm{\sourcemixtureempdistrib}{\targetempdistrib}{\fclass}
        +
        \sup_{\f\in\fclass}
        \left|
            \E_{\targetempdistrib} \f(\point)
            -
            \E_{\targetdistrib} \f(\point)
        \right|
        ,
    \end{align*}
    where in the first inequality, we applied the triangle inequality,
    the second inquality, we split the supremum terms, and
    and in the last line, we applied \Cref{def-ipm}.

    The term
    $
        \sup_{\f\in\fclass}
        \left|
            \E_{\targetempdistrib} \f(\point)
            -
            \E_{\targetdistrib} \f(\point)
        \right|
    $
    can be bounded in a variety of ways.
    Here, we use a standard bound via the empirical Rademacher complexity
    (\Cref{lemma-symmetrization}) to yield the result.
\end{proof}

The bound in \Cref{thm-second}
involves purely empirical quantities, i.e.,
the empirical IPM and empirical Rademacher complexity.
Note that only the empirical IPM in the first
term involves the $\alpha$-weights.

Since $\alpha\in\Delta^{\numsources-1}$, it follows that
\begin{align*}
    \ipm{\sourcemixtureempdistrib}{\targetempdistrib}{\fclass}
    & =
    \sup_{\f\in\fclass} \left|
        \sum_{\sourceindex=1}^\numsources
        \alpha_\sourceindex
        \E_{\sourceempdistrib}
        \f(\point)
        -
        \E_{\targetempdistrib}
        \f(\point)
    \right|
    \\
    & \leq
    \sup_{\f\in\fclass}
    \sum_{\sourceindex=1}^\numsources
    \alpha_\sourceindex
    \left|
        \E_{\sourceempdistrib}
        \f(\point)
        -
        \E_{\targetempdistrib}
        \f(\point)
    \right|
    \\
    & \leq
    \sum_{\sourceindex=1}^\numsources
    \alpha_\sourceindex
    \ipmdef{\sourceempdistrib}{\targetempdistrib}
    \\ &
    =
    \sum_{\sourceindex=1}^\numsources
    \alpha_\sourceindex
    \ipm{\sourceempdistrib}{\targetempdistrib}{\fclass}
    .
\end{align*}
This immediately yields the corollary:
\begin{corollary}
    \label{corr}
    Assume the conditions of \Cref{thm-second} hold.
    Then with probability at least $1-\epsilon$,
\begin{align}
    \ipm{\sourcemixtureempdistrib}{\targetdistrib}{\fclass}
    \leq
    \sum_{\sourceindex=1}^\numsources
    \alpha_\sourceindex
    \ipm{\sourceempdistrib}{\targetempdistrib}{\fclass}
    +
    2 \mathcal{R}(\fclass|\point_1,\dots,\point_{\size{\targetindex}})
    +
    3\sqrt{\frac{(b-a)^2\log(2/\epsilon)}{2\size{\targetindex}}}
    .
\end{align}
\end{corollary}
While the weighted empirical IPM in \Cref{corr} results in a looser bound,
it leads to an even simpler and computationally cheaper weight selection rule,
which may be sufficient for some problems;
we discuss this further in \Cref{sec-algorithm}.

\Cref{corr} can be interpreted
as an empirical version of the bound in
\citet[Theorem~5.2]{zhang2013arxiv}
for the function class $\G$ defined in \Cref{eq-erm};
in
\citet[Theorem~5.2]{zhang2013arxiv},
the upper bound on
$\ipm{\sourcemixtureempdistrib}{\targetdistrib}{\fclass}$
includes a population IPM with respect to $\G$
and a weighted sum of expected Rademacher complexity terms on the source domains.

\begin{figure*}[t]
    \centering
    \includegraphics[scale=0.365]{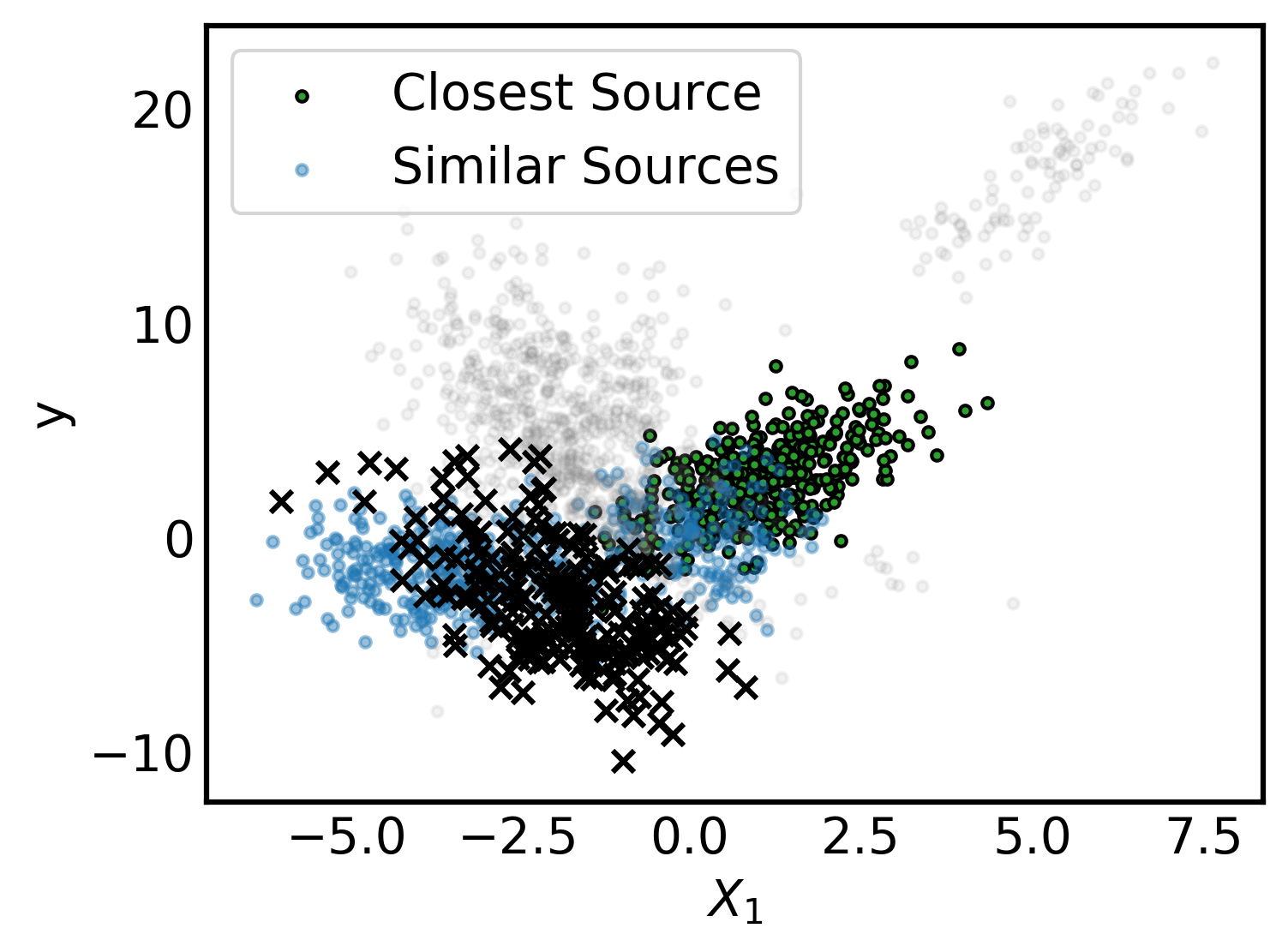}
    \includegraphics[scale=0.365]{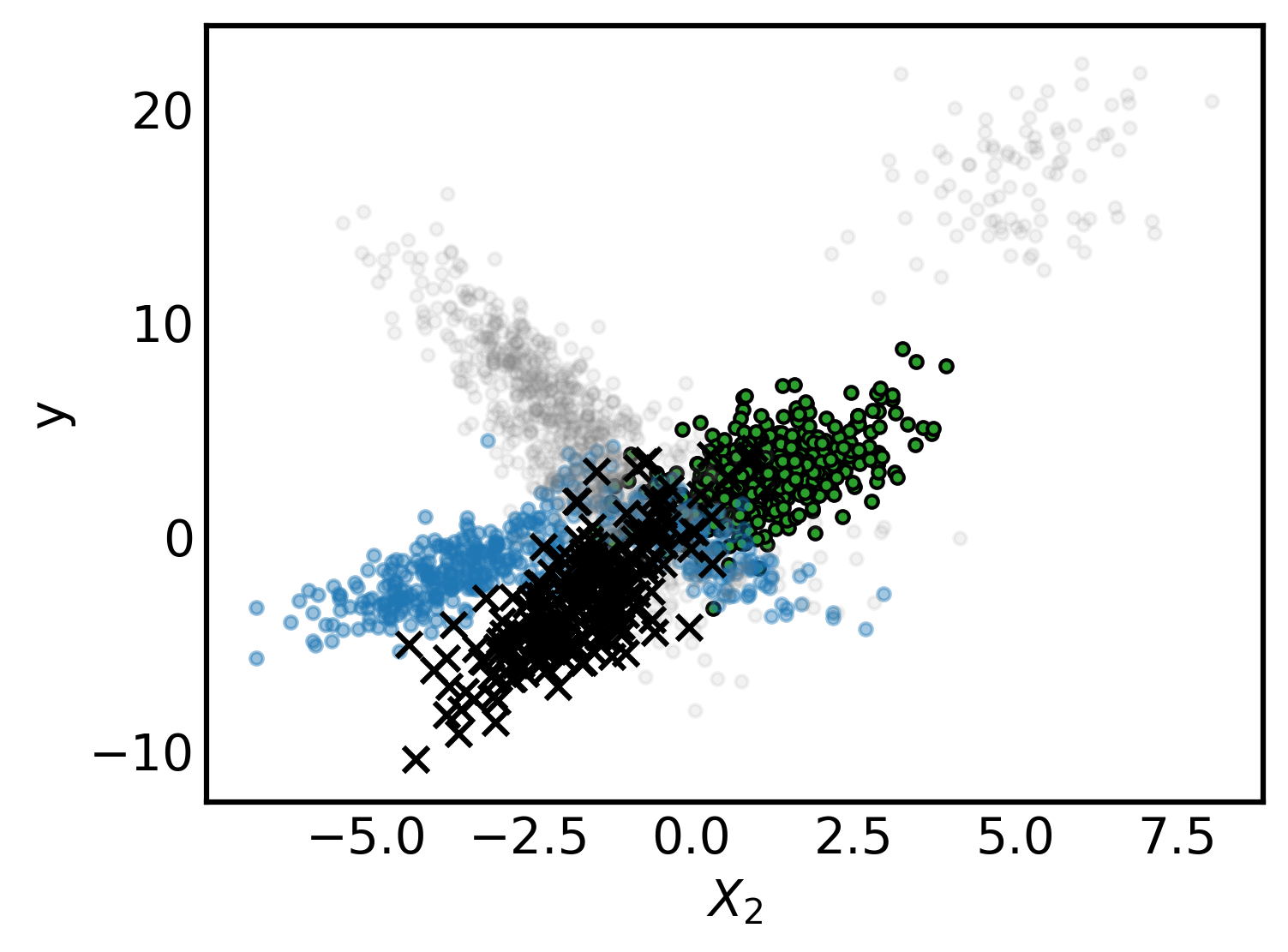}
    \includegraphics[scale=0.365]{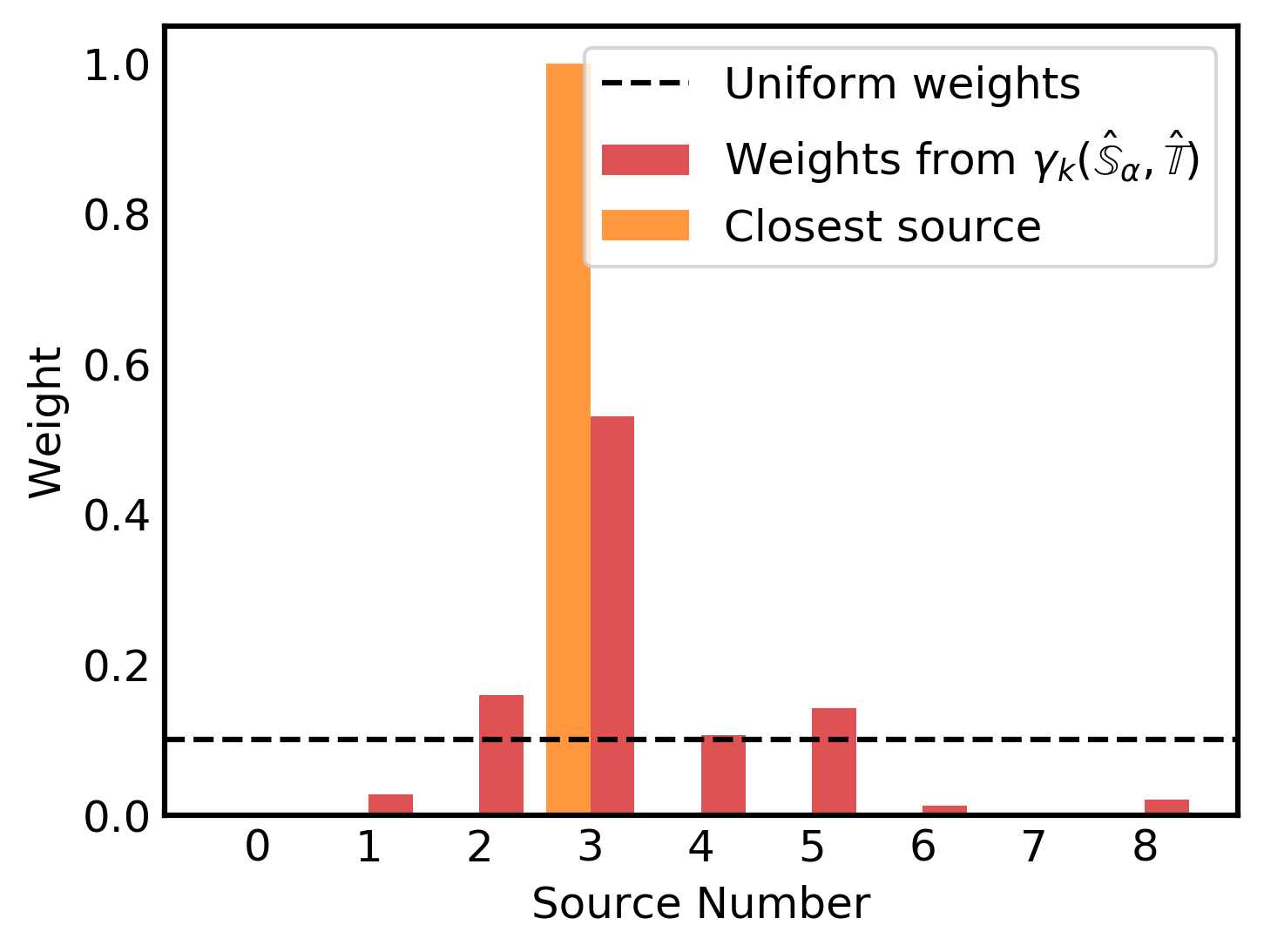}
    \caption{
        2-dimensional data sampled from isotropic Gaussian distributions.
        \textbf{Left, Middle:}
        The black x's denote a target task,
        and the green points denote the closest source,
        which receives full weight
        when minimizing a bound based on \Cref{corr}.
        The blue sources denote other sources that
        receive greater than $1/J$ weight
        when minimizing the bound based on
        \Cref{thm-second},
        where $J$ is the number of sources, and
        the gray black circles denote the remaining sources that have weight
        less than $1/J$.
        \textbf{Right:}
        The respective weightings
        from minimizing the bounds based on
        \Cref{thm-second}
        and
        \Cref{corr}.
    }
    \label{fig-distances}
\end{figure*}

In \Cref{fig-distances}, we show an example of a 2-dimensional regression task
with 9 source tasks, each generated from an isotropic Gaussian distribution.
In the two leftmost figures, the dark black points represent the
target task, and the green points represent the most similar task.
Minimizing a bound based on \Cref{corr} is equivalent to
putting all weight on the closest task.
By constrast, minimizing the bound based on \Cref{thm-second},
allows for finding the best mixture of source tasks such that this mixture
is close to the target; the additional source tasks with weight greater than $1/J$
are highlighted in blue (in addition to the source in green).
The weights found by minimizing  a kernel distance (see \Cref{sec-algorithm})
are plotted in the rightmost graph.


\section{Algorithm for weight selection using empirical kernel distances}
\label{sec-algorithm}

The upper bounds given in \Cref{thm-second} and \Cref{corr}
lead naturally to an algorithm for computing
the weights by minimizing the bound.
However,
optimizing the upper bound of \Cref{thm-second} or \Cref{corr} requires computing the IPM
$
    \ipm{\sourcemixtureempdistrib}{\targetempdistrib}{\fclass}
$
or IPMs
$
    \{ \ipm{\sourceempdistrib}{\targetempdistrib}{\fclass} \}_{\sourceindex=1}^{\numsources}
$,
which are in general not computable for arbitary
function classes $\G$ \citep{sriperumbudur2012empirical}.
Thus, the goal is to compute a surrogate distance
that provides an upper bound on the empirical IPMs.

\subsection{The kernel distance}

One candidate for such a surrogate distance is the kernel distance,
which is an integral probability metric defined with respect to
the class of functions given by
the unit ball of a reproducing kernel Hilbert space (RKHS), i.e.,
$\G_\text{RKHS} := \{g: \norm{g}_{\mathcal{K}_k} \leq 1\}$,
where $\mathcal{K}_k$ is a Hilbert space associated with
a reproducing kernel $k: \Z \times \Z \rightarrow \reals$
and $\norm{\cdot}_{\mathcal{K}_k}$ is the norm induced by
the inner product on $\mathcal{K}_k$.
That is, the Hilbert space $\mathcal{K}_k$
associated with a reproducing kernel $k$
has the properties that (1) for all
$z \in \Z$, $k(\cdot, z) \in \mathcal{K}_k$
and (2) for all $z \in \Z$ and for all functions $g \in \mathcal{K}_k$,
$g(z) = \langle g, k(\cdot, z)\rangle_{\mathcal{K}_k}$.

Let $\ipm{\mathbb{P}}{\mathbb{Q}}{\fclass_\text{RKHS}}$
denote the kernel distance with respect to
the probability distributions $\mathbb{P}$ and $\mathbb{Q}$.
In order to upper bound the IPMs defined with respect to $\G$,
we need to find an
RKHS ball $\fclass_\text{RKHS}$
associated with a kernel $k$
such that the function class is contained in the
RKHS ball, i.e., $\fclass\subseteq\fclass_\text{RKHS}$.
Then respective IPMs can then be bounded as
\begin{align}
    \label{eq-bound1}
    \ipm{\sourcemixtureempdistrib}{\targetempdistrib}{\fclass}
    \le
    \ipm{\sourcemixtureempdistrib}{\targetempdistrib}{\fclass_\text{RKHS}}
    :=
    \ipm{\sourcemixtureempdistrib}{\targetempdistrib}{k}
\end{align}
and for $1\le\sourceindex\le\numsources$,
\begin{align}
    \label{eq-bound2}
    \ipm{\sourceempdistrib}{\targetempdistrib}{\fclass}
    \le
    \ipm{\sourceempdistrib}{\targetempdistrib}{\fclass_\text{RKHS}}
    :=
    \ipm{\sourceempdistrib}{\targetempdistrib}{k},
\end{align}
where $\ipm{\cdot}{\cdot}{k}$
is the empirical kernel distance, or maximum mean discrepancy.

The empirical kernel distance
between the $\alpha$-weighted source distribution
and the target distribution
$\ipm{\sourcemixtureempdistrib}{\targetempdistrib}{k}$
can be easily computed
(\citet[Theorem~2.4]{sriperumbudur2012empirical})
as
\begin{align}
    \label{eq-kernel-dist-computation}
    \ipm{\sourcemixtureempdistrib}{\targetempdistrib}{k}
    =
    \sqrt{v_\alpha^\top \, K_\numsources \, v_\alpha},
    \qquad
    v_\alpha &:= \left[\frac{\alpha_1}{N^{(1)}},\ldots, \frac{\alpha_\numsources}{N^{(J)}},
    \frac{-1}{N^{(T)}}\right]^\top \in \reals^{J+1},
\end{align}
where
$K_J \in \reals^{(J+1)\times (J+1)}$ is
a kernel gram matrix between tasks,
with $[K_J]_{j,j'} = \sum_{i,i'} k(z_i^{(j)}, z_{i'}^{(j')})$.
The empirical kernel distance for a single
source and target distribution
can then be computed as
$\ipm{\sourceempdistrib}{\targetempdistrib}{k}
= \ipm{\hat{\mathbb{S}}_{e_j}}{\targetempdistrib}{k},
$
where $e_j \in \Delta^{J-1}$ is the $j$-th standard basis vector.

Minimizing a bound based on the kernel distance
in  \Cref{eq-kernel-dist-computation}
involves solving a quadratic program with simplex constraints,
which has time complexity $O(J^3)$ when
the gram matrix $K_J$ is positive definite.
Finally, an even simpler approach than
optimizing a bound based on \Cref{thm-second} (via \Cref{eq-kernel-dist-computation})
is optimizing a bound based on \Cref{corr},
which only requires computing each
the distances $\{\ipm{\sourceempdistrib}{\targetempdistrib}{k}\}_{j=1}^J$
once and reporting the minimum distance.

\subsection{Upper bounds on the IPM for linear basis models}
\label{ssec-examples}

We have established that if
$\fclass\subseteq\fclass_\text{RKHS}$,
then the IPM $\gamma_\G$ with respect to $\G$ can be upper bounded
with a computable empirical kernel distance $\gamma_{k}$ with respect to the
RKHS ball $\G_{\text{RKHS}}$,
as in \Cref{eq-bound1} and \Cref{eq-bound2}.
We now show how to construct a
class of functions $\fclass_\text{RKHS}$ such that
$\fclass\subseteq\fclass_\text{RKHS}$
for regression and binary classification settings;
concretely, we consider linear basis models
with square loss and hinge loss functions, respectively.

Let $\modelbasis:\lininputspace\rightarrow\reals^d$ denote a basis function,
and consider the class of linear basis function models
composed with a loss $\loss$,
\[
    \fclass^\loss
    :=
    \{
        \f( (\inputpoint, \outputpoint) )
        =
        \loss(
            \outputpoint,
            \lincoeff^\top \modelbasis(\inputpoint;\modelbasisparam)
        )
        :
        \lincoeff\in\lincoeffspace,
        \modelbasisparam\in\modelbasisparamspace
    \},
\]
where $\lincoeffspace \subset \reals^d$ denotes a constraint set
and $\Theta$ denotes the parameter space for the basis function $\psi$.
We consider selecting kernels for the class of functions
$\G^\ell$
such that $\fclass^\loss\subseteq\fclass_\text{RKHS}$,
where $\fclass_\text{RKHS}$ is a RKHS ball associated with the kernel.
To do so, we define a feature map $\phi$
mapping $\reals^{d+1}$ to a Euclidean feature space,
and define $\G_{\text{RKHS}}$ to be a ball of an RKHS $\K_k$ constructed from
the kernel $k(z,z') = \langle \phi(\psi(x),y), \phi(\psi(x'), y')\rangle$.

In the following, let
$\point$ denote a point $(\modelbasis(\inputpoint),\outputpoint)$,
and let
$\text{vec}(\cdot)$ denote the vectorization operator.
First we consider the class of functions $\G^\ell$
when $\ell$ is a square loss function.
\begin{lemma}[Square loss]
Let $\lincoeffspace=\{\lincoeff\in\mathbb{R}^d : \Vert \lincoeff \Vert_2 \le 1\}$.
For $\loss( \outputpoint, \outputpoint') = \frac{1}{2}(\outputpoint - \outputpoint')^2$,
construct an RKHS from the feature map $\phi : \mathbb{R}^{d+1} \rightarrow \mathbb{R}^{d^2+d+1}$,
\[
\kernelbasis( (\modelbasis(\inputpoint),\outputpoint) )
=
\begin{pmatrix}
    \text{vec}(\modelbasis(\inputpoint) \modelbasis(\inputpoint)^\top) \\
    \sqrt{2} \outputpoint \modelbasis(\inputpoint) \\
    \outputpoint^2
\end{pmatrix}
.
\]
Then, $\fclass^\loss\subseteq\fclass_\text{RKHS}$ for the kernel $k$ associated with the feature map $\kernelbasis$.
\end{lemma}
\begin{proof}
Fix $\lincoeff\in\lincoeffspace$ and
let $a_1=\frac{1}{2},z_1=(-\lincoeff,1)$.
Let $g \in \G^\ell$.
Then
\begin{align}
\label{eq-square-loss-function}
g(z) &=
\loss( \outputpoint, \lincoeff^\top \modelbasis(\inputpoint))
\nonumber \\
    &=
\frac{1}{2}(\lincoeff^\top \modelbasis(\inputpoint) - \outputpoint)^2
\nonumber \\
    &=
\frac{1}{2}
(
    \text{vec}(\psi(\inputpoint) \psi(\inputpoint)^\top)^\top
    \text{vec}(\lincoeff \lincoeff^\top)
    - 2 \outputpoint \modelbasis(\inputpoint)^\top \lincoeff
    + \outputpoint^2
)
\nonumber \\
&=
a_1\phi(\point)^\top \phi(\point_1)
= a_1 k(\point, \point_1)
\in \mathcal{K}_k.
\end{align}
Applying \Cref{eq-square-loss-function}
    and Property~(2) of the RKHS,
    $g$ has bounded norm:
\begin{align*}
    \Vert \f \Vert_{\mathcal{K}_k}^2
    &=
    \langle \f, \f \rangle_{\K_k}
    =
    a_1 \langle \f, k(\cdot,z_1) \rangle_{\K_k}
    =
    a_1^2 k(\point_1,\point_1)
    =
    a_1^2 (
        \Vert \lincoeff \Vert_2^2 + 2 \Vert \lincoeff \Vert_2 + 1
    )
    \le 1,
\end{align*}
where the inequality follows from
the assumption that $\norm{w}_2 \leq 1$.
Thus,
$\fclass^\loss\subseteq\fclass_\text{RKHS}$.
\end{proof}

Now we consider $\G^\ell$ where $\ell$
is a hinge loss function, with constraints
on the domain and parameter spaces.
This allows us to, e.g., utilize a penalized SVM with sufficiently small penalty $C$ on the solution norm, i.e., $\Vert\lincoeff\Vert^2 \le C$.
\begin{lemma}[Hinge loss]
\label{lemma-hinge}
Let
$\lincoeffspace=\{\lincoeff\in\mathbb{R}^d : \Vert \lincoeff \Vert_2 \le 1\}$,
$\modelbasisparamspace=\{\modelbasisparam : \Vert \modelbasis(\inputpoint;\modelbasisparam) \Vert_2 \le 1\}$,
$\linoutputspace=[-1,1]$.
Under the constraints on the input and output spaces,
\[\loss(\outputpoint,\outputpoint')=\text{max}(1-\outputpoint \modelbasis(\inputpoint)^\top
\lincoeff,0)=1-\outputpoint \modelbasis(\inputpoint)^\top \lincoeff.\]
Construct an RKHS from the feature map $\kernelbasis: \mathbb{R}^{d+1} \rightarrow \mathbb{R}^{d+1}$,
\[
\kernelbasis( (\modelbasis(\inputpoint),\outputpoint) )
=
\begin{pmatrix}
    \outputpoint \modelbasis(\inputpoint) \\
    1
\end{pmatrix}
.
\]
Then, $\fclass^\loss\subseteq\fclass_\text{RKHS}$ for the kernel $k$ associated with the feature map $\kernelbasis$.
\end{lemma}
\begin{proof}
Fix $\lincoeff\in\lincoeffspace$ and let $\point_1=(-\lincoeff,1)$.
Let $g \in \G^\ell$.
Then, $g$ is an element of the RKHS $\K_k$, i.e.,
\begin{align*}
    \f(\point) =
    \loss( \outputpoint, \lincoeff^\top \modelbasis(\inputpoint))
    =
    1-\outputpoint \modelbasis(\inputpoint)^\top \lincoeff
    =
    k(\point,\point_1)
    \in \K_k,
\end{align*}
which, along with Property (2) of the RKHS,
implies that $g$ has bounded norm:
    \[
    \Vert \f \Vert_{\K_k}^2
    =
    \langle \f, \f \rangle_{\K_k}
    =
    \langle \f, k(\cdot,z_1) \rangle_{\K_k}
    =
    k(\point_1,\point_1)
    =
    \Vert \lincoeff \Vert_2^2 + 1
    \le 2,
\]
where we applied the assumption that
$\norm{w}_2 \leq 1$.
Thus,
$\fclass^\loss\subseteq\fclass_\text{RKHS}$.
\end{proof}

Note that in \Cref{lemma-hinge}, $\G_{\text{RKHS}}$ is a $\sqrt{2}$-RKHS ball;
the extra constant factor only scales the kernel distance computation
and therefore does not affect the computation of the weights.

Thus, since the upper bounds on the empirical IPM
in \Cref{eq-bound1} and \Cref{eq-bound2}
hold for linear basis functions with
square and hinge loss,
we can apply a weight minimization
algorithm based on minimizing the kernel distance,
instead of the IPMs in \Cref{thm-second}
and \Cref{corr}.

This construction encodes a natural notion for task similarity:
when the kernel distance between two tasks is relatively small, this implies that the model class
cannot distinguish between these tasks with respect to the associated loss function.
Hence, a model learned on one task should perform similarly on the other task.

\subsection{Weighted meta-learning for linear basis models}

The examples in \Cref{ssec-examples}
examine classes of linear basis functions
composed with a loss $\G^\ell$
without explicitly considering an adaptation
function $U$.
\citet{finn19a}
summarizes sufficient conditions of under which the projection in
\Cref{eq:maml-u-function}
is equivalent to a contraction, ensuring that model updates during training
remain within $\fclass^\loss$.
More generally, a projection step back into $\fclass^\loss$ can be utilized
during optimization.

\Cref{algorithm-alpha} summarizes the meta-learning procedure used
learn the $\alpha$ weight values and an initial model.
Note that in an adaptive basis setup, steps 2--4 are iterated, since
selecing $g$ changes the basis function $\psi$.

\begin{algorithm}[t]
\centering
\caption{Meta-training procedure for $\alpha$-meta-learning}
\label{algorithm-alpha}
    \begin{algorithmic}[1]
        \STATE \textbf{Input:}
        kernel $k$, source tasks $\{Z_j\}_{j=1}^J$, target task $Z^T$
       \STATE{
           Compute empirical kernel distance
        $\ipm{\sourcemixtureempdistrib}{\targetempdistrib}{k} =
        \sqrt{v_\alpha^\top \, K_\numsources \, v_\alpha}$
        }
       \STATE{Compute $\hat\alpha := \arg\min_{\alpha \in \Delta^{J-1}}
        \ipm{\sourcemixtureempdistrib}{\targetempdistrib}{k}$
       }
       \STATE{Learn initial model by minimizing
             $\sum_{j=1}^J \hat\alpha_j \E_{\sourceempdistrib} g(z)$
        }
        \STATE \textbf{Output:} weights $\hat\alpha$ and initial model $\hat g$
   \end{algorithmic}
\end{algorithm}


\section{Experiments}
\label{sec-experiments}

We present several regression examples on synthetic and real data tasks
that use $\alpha$-weighted meta-learning and compare to
the uniformly-weighted setting, which recovers algorithms such as
MAML and joint training.
Additional experimental details can be found in \Cref{appendi-experiments}.

\subsection{Synthetic linear regression}

First we examine a 1-dimensional linear regression setting.
We generated 9 source tasks and 1 target task as follows.
The task sizes were generated according to a multinomial
distribution with a uniform prior on the multinomial parameter.
For each source, the covariates
were generated from a gaussian with mean $\mu_j$ and variance 1,
where $\mu_j \sim \text{uniform}(-5,5)$.
The slope of the $j$-th task was set to $2\mu_j$,
and the response of the $j$-th source was then drawn according to
$y_j \sim 2\mu_j + \epsilon$, where $\epsilon \sim \text{N}(0,1)$.

We note that for weighted MAML and weighted ERM,
an analytical solution to the meta-objective can be computed, see \Cref{ssec-maml-soln} for a derivation.
Thus, the analytical solution is used to compute an initialization,
and we compare the resulting initializations from $\alpha$-weighted meta-learning
and uniform weighting.
All MAML solutions were computed with $\eta = 0.0001$ step size.

\begin{figure*}
    \centering
    \includegraphics[scale=0.50]{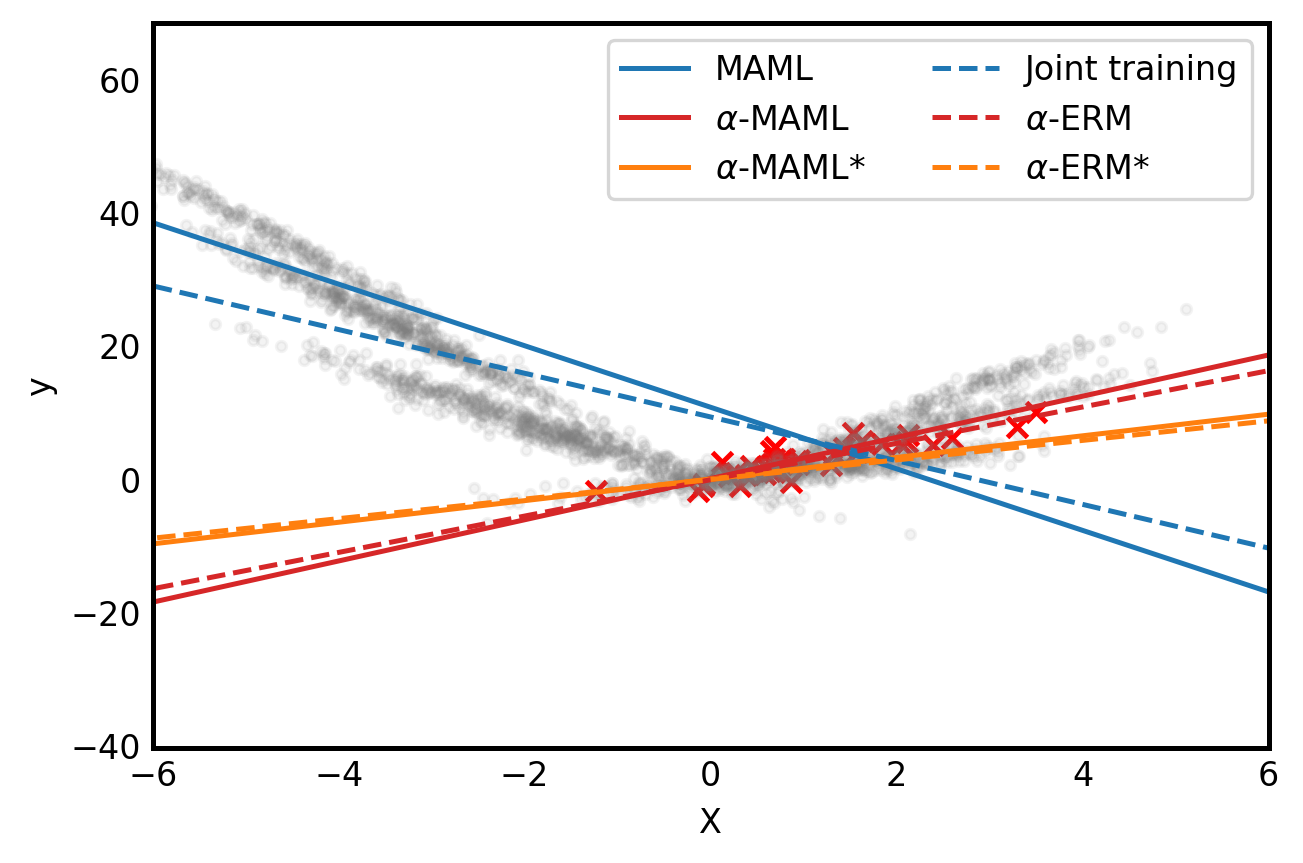}
    \includegraphics[scale=0.50]{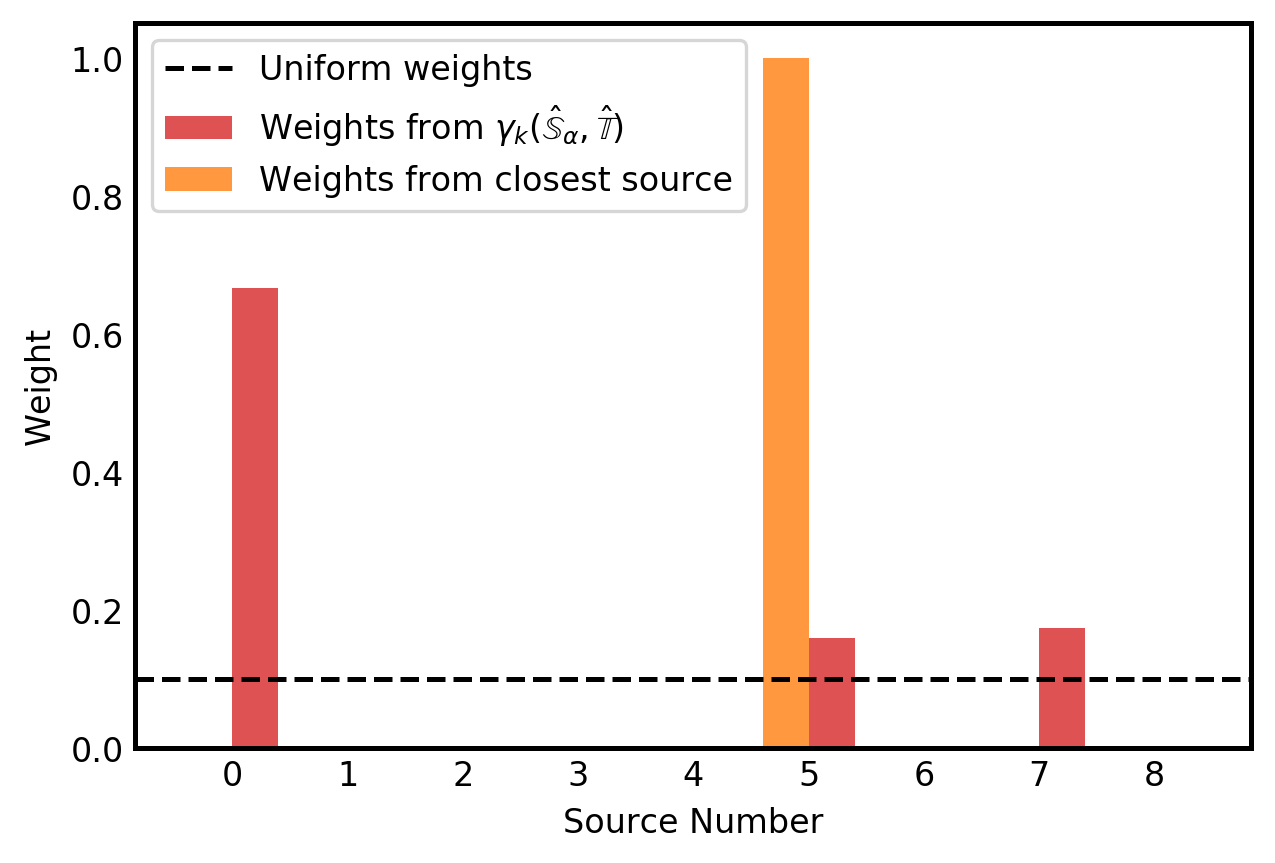}
    \caption{
        \textbf{Left:}
        Linear regression with MAML, joint training, $\alpha$-MAML, and $\alpha$-ERM solutions.
        Red x's denote the target task, and gray points denote the 9 source tasks.
        \textbf{Right:}
        The various weightings obtained from
        uniform weighting, $\alpha$-weighting according to the kernel IPM between the $\alpha$-mixture
        of sources and target, or
        weighting only the closest source.
    }
    \label{fig-1d-regression}
\end{figure*}

In \Cref{fig-1d-regression}, we plot the initializations obtained from
each method.
Here $\alpha$-MAML and $\alpha$-ERM denote the initializations
from minimizing the bound with the kernel distance, whereas
$\alpha$-MAML$^*$ and $\alpha$-ERM$^*$
denote the initializations obtained from placing all weight on the closest source.
The kernel distance was computed using
20 target training examples (denoted by red points).
Lines denote inferred hypotheses using uniform, $\alpha$, and closest source weightings.
We observe that the uniformly-weighted intializations (blue)
correspond to an average model learned from all the sources,
whereas unequally weighted sources (red, orange) are able to
use the task similarity to better represent the target task.
The weights are shown in the bottom plot in \Cref{fig-1d-regression},
where the weights  obtained from minimizing the kernel distance
$\ipm{\sourcemixtureempdistrib}{\targetempdistrib}{k}$
place weight on 3 sources.

\subsection{Sine regression with $\alpha$-MAML}

We generated synthetic sine wave tasks as follows.
The target task was assigned a fixed amplitude of 6 with
a small number of samples (5, 10, 20)
for training and fast adaptation,
and 100 data points were randomly sampled from the target task for evaluation.
For the source tasks,
the amplitudes were drawn according to a $\text{gamma}(1,2)$ distribution.
For both target and source tasks, the phase parameter was drawn uniformly from
$(0, \pi)$, as in the setup in \citet{finn2017model}.
From each source task, 40 samples were drawn, where the $x$ values were sampled
uniformly from $(-5,5)$.

Following \citet{finn2017model}, we used a
fully-connected neural network with 2 hidden layers of size 40 with
ReLU non-linearities.
For all experiments, Adam was used as the meta-optimizer with an inner-loop learning rate of
$0.01$ and an outer-loop learning rate of $0.001$.
In each iteration of weighted MAML, we
sample a mini-batch of $T$ tasks,
compute embeddings for $T$ tasks and the target task,
compute weights by optimizing the bound (and computing kernel
    distances using the computed embeddings of the sources and target),
    compute weighted loss using the optimal weights, and
    lastly,
update the model parameters.
We used the same as above for MAML  but with uniformly weighted source tasks.
For both MAML and weighted MAML, the mini-batch size was set to $T=100$ tasks.

\begin{figure*}[t]
    \centering
    \includegraphics[scale=0.50]{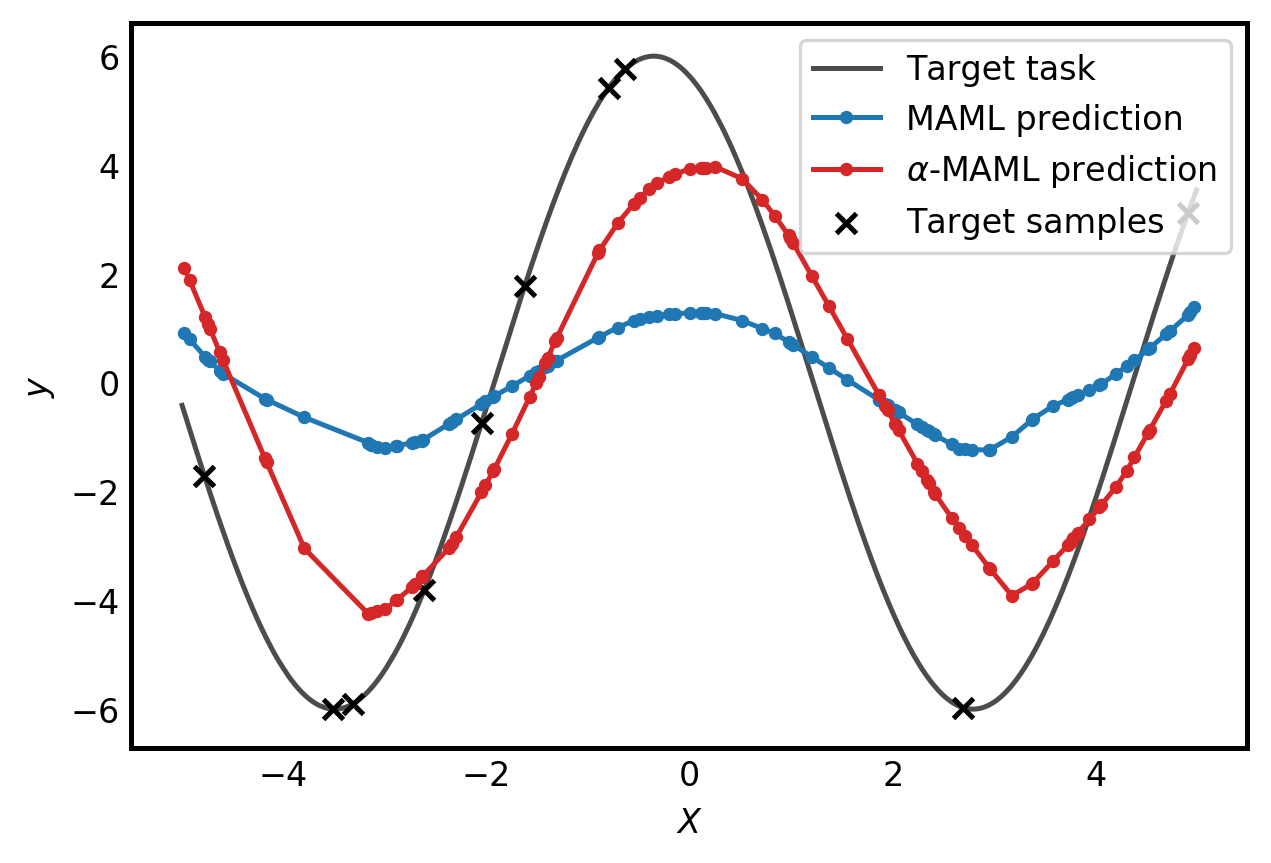}
    \includegraphics[scale=0.50]{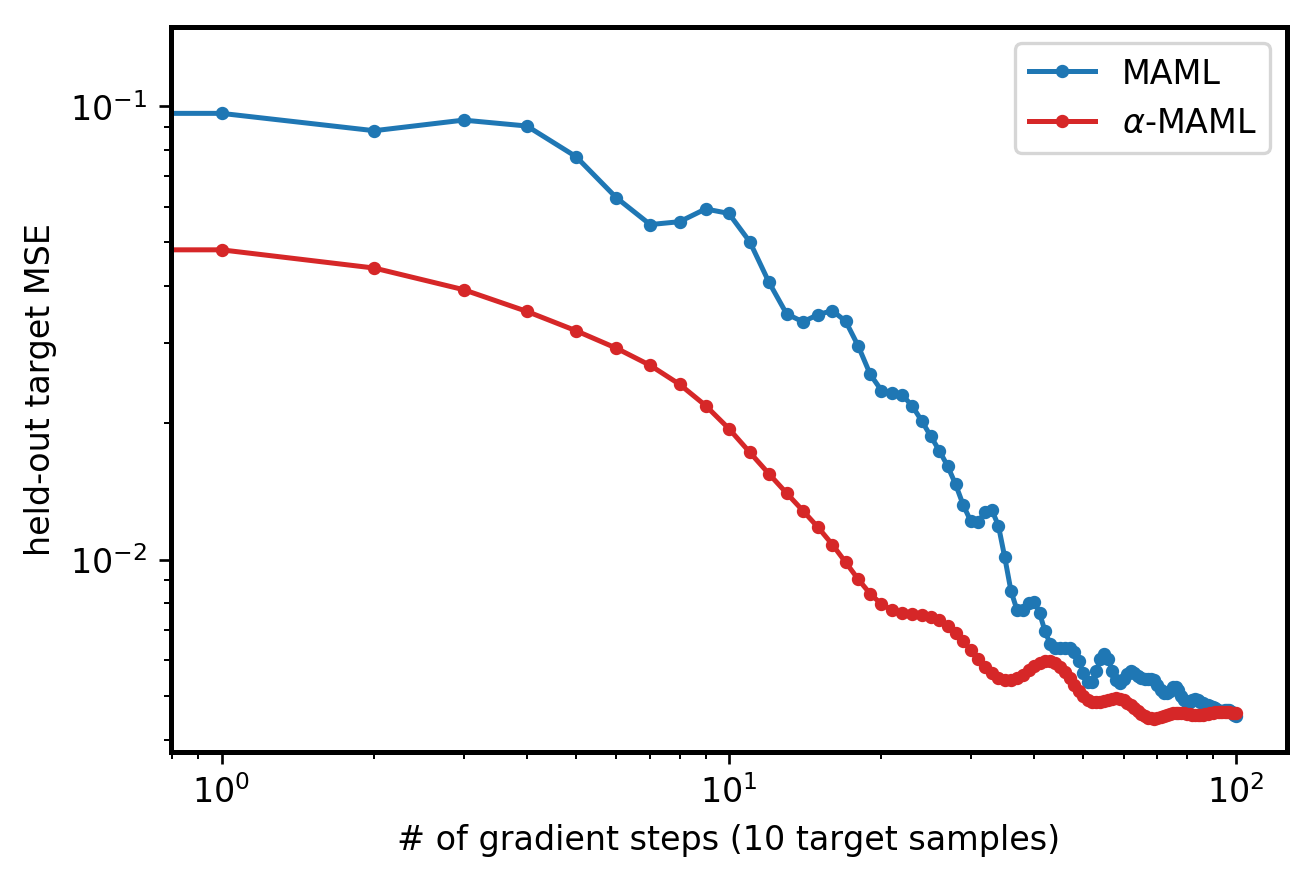}
    \caption{
        Sine wave regression with 10 labeled target examples.
        \textbf{Left:}
        Predictions after training MAML and weighted MAML for 10,000
        meta-iterations.
        \textbf{Right:}
        Held-out target mean squared error after $L$ gradient steps of
        fast adaptation.
    }
    \label{fig-sine-regression}
\end{figure*}

\Cref{fig-sine-regression} shows
one target task where 10 training samples
(denoted by black points)
are drawn from the target (denoted by the solid black curve).
The red and blue curves denote the resulting predictions
from the learned intializations of uniformly-weighted MAML and $\alpha$-weighted MAML.
In the bottom plot,
the intializations are adapted to the 10 target samples.
In this plot, we observe that only after a larger number of gradient
steps ($\sim$100) is the uniform weighting able to achieve a comparable
mean squared error on the held-out target samples as the $\alpha$-weighted
initialization.

In \Cref{table-sine}, we report the average RMSE for each method before and after fast adaptation
for MAML and ERM with 1) uniform weights, 2) $\alpha$-weights
(\Cref{algorithm-alpha}), and 3) threshold weights (i.e., closest source selection).
In the table, we see that the predictions
from the initializations are fairly close
for all methods, with the threshold method
and $\alpha$-MAML achieving lower RMSE on the
predicted values than uniformly-weighted MAML on average.
On average, the $\alpha$-MAML is able to adapt better in 10 gradient steps
than the uniformly and single-source threshold MAML intializations,
and the threshold method still is competitive for
fast adaptation,
especially relative to uniformly weighted MAML.

\begin{table}
   \centering
    \caption{
        RMSE of sine wave predictions using
        (1) the initial meta-model and (2) after 10
        gradient steps (denoted by $\dagger$) for 5-shot, 10-shot, and 20-shot target training
        scenarios, averaged over 4 random trials.
    }
    \label{table-sine}
    \begin{tabular}{cccc}
                      \toprule
                      & 5-shot & 10-shot & 20-shot \\
                      \midrule
         MAML         & $3.90 \pm 0.85$ &  $3.57 \pm 0.66$ & $4.11 \pm 0.94$   \\
        $\alpha$-MAML & $3.21 \pm 1.12$ &  $2.93 \pm 0.75$ & $3.05 \pm 1.09$ \\
        Threshold     & $2.83 \pm 1.04$ &  $3.17 \pm 1.05$ & $3.26 \pm 1.08$ \\
        \midrule
         MAML$^\dagger$         & $4.24 \pm 1.00$ &  $1.90 \pm 0.26$ & $2.06 \pm 0.39$   \\
        $\alpha$-MAML$^\dagger$ & $2.65 \pm 1.34$ &  $1.68 \pm 0.77$ & $1.67 \pm 0.75$   \\
         Threshold$^\dagger$    & $2.35 \pm 1.75$ &  $2.01 \pm 0.83$ & $2.01 \pm 0.88$   \\
              \bottomrule
    \end{tabular}
\end{table}

\subsection{Weighted meta-learning for real data tasks}

\paragraph{Multi-dimensional linear regression.}

We examined two multi-dimensional regression data sets.
The first uses the diabetes data set studied by
\citet{efron2004least}, which contains 10 covariates.
The goal is to predict a real-valued response
that measures disease progression one year after baseline.
We split the data set into separate source tasks by grouping on age,
leading to a total of 6 source tasks,
using the remaining covariates in each source.
We picked a separate age group for the target task,
using 20 target samples for computing the kernel distance
and the remaining target samples were used for testing.

The second data set
is the Boston house prices data of \citet{harrison1978hedonic},
which includes 13 covariates,
and the response variable is
the median value of owner-occupied homes.
To form sources, we grouped on the attribute
age, and separated the full data into
6 source tasks, where each source contained a group of 50 ages,
and the remaining 12 covariates were used in each source.
The target task contained 30 target samples for training,
and the remaining samples were used for testing.

The root mean squared error (RMSE) of each of the initializations obtained are
presented in \Cref{table-mult-regression},
where all MAML-related computations used $\eta=0.0001$ for the step size.
This is a setting where predicting on
only the target training samples performs quite poorly and
using the source data sets improves performance for this particular target task.
Furthermore, weighting sources by kernel distance
seems to also improve prediction error.
We found that typically, similar age ranges were upweighted more than
further away age groups.

\begin{table}
\centering
\caption{RMSE of initializations for linear regression on sources and  target
using 20 labeled target training examples to
compute the weights (before fast adaptation).
}
\label{table-mult-regression}
\begin{tabular}{ccc}
           \toprule
    & Diabetes & Boston  \\
    \midrule
    MAML          & 50.44 &  3.64   \\
    $\alpha$-MAML & 49.36 &   3.59  \\
    \midrule
    Joint training & 50.31 & 3.58     \\
    $\alpha$-ERM & 49.24 & 3.32     \\
    \midrule
    target & 92.31 & 15.47          \\
    \bottomrule
\end{tabular}
\end{table}

\paragraph{Basis linear regression.}

Next we examined the sales data set studied by
\citet{tan2014time}.
The data set consists of a collection of products with sales information over 52 weeks.
We included the 300 products as source tasks, and used a single product as the target task.
For the target task, we used the first 10 weeks as labeled target training data,
and the last 42 weeks as test data for evaluation.
We computed random Fourier features \citep{rahimi2008random} for the weeks
and used these features when computing the $\alpha$-weights in the kernel distance.

In \Cref{table-sales},
we report the RMSE  on held-out target data for
target tasks with 5 and 10 weeks of data,
i.e., 5- and 10-shot target tasks,
averaged over 20 different product target tasks.
In this example, the table shows that
predicting on the target data alone performs
very poorly but that meta-learning helps improve
performance.

Here $\alpha$-MAML and $\alpha$-ERM are the methods used in
\Cref{algorithm-alpha} for MAML and ERM, respectively,
whereas thresh-MAML and thresh-ERM correspond
to the threshold method that weights the closest source only.
In this setting,
both weighted methods, i.e., $\alpha$-based and thresh-based meta-learning,
outperform the uniformly-weighted methods.

\begin{table}
\centering
\caption{
    RMSE of sales data for 5-shot and 10-shot target training sample sizes,
    where the remaining data was used for evaluation.
    Mean and standard deviation computed over 20 target tasks.
}
    \label{table-sales}
\begin{tabular}{ccc}
        \toprule
    &  5-shot & 10-shot \\
    \midrule
MAML    & $ 12.86\pm 3.79$    &  $12.69 \pm    3.69   $  
\\
$\alpha$-MAML & $ 2.43\pm 2.09  $    &  $2.41 \pm	1.92   $  
\\
thresh-MAML  & $ 2.53\pm 1.94  $    &  $2.50 \pm	2.03   $  
\\
\midrule
ERM	    & $ 12.09\pm 3.53$    &  $11.92 \pm	3.43   $  
\\
$\alpha$-ERM  & $ 2.52\pm 2.21$    &  $2.50 \pm	2.04   $  
\\
thresh-ERM   & $ 2.45\pm 1.81$    &  $2.45 \pm	1.97   $  
\\
\midrule
target  & $ 83.63\pm 104.07 $ &  $209.06 \pm  378.88 $   
\\
\bottomrule
\end{tabular}
\end{table}

\begin{figure}
   \centering
    \includegraphics[scale=0.5]{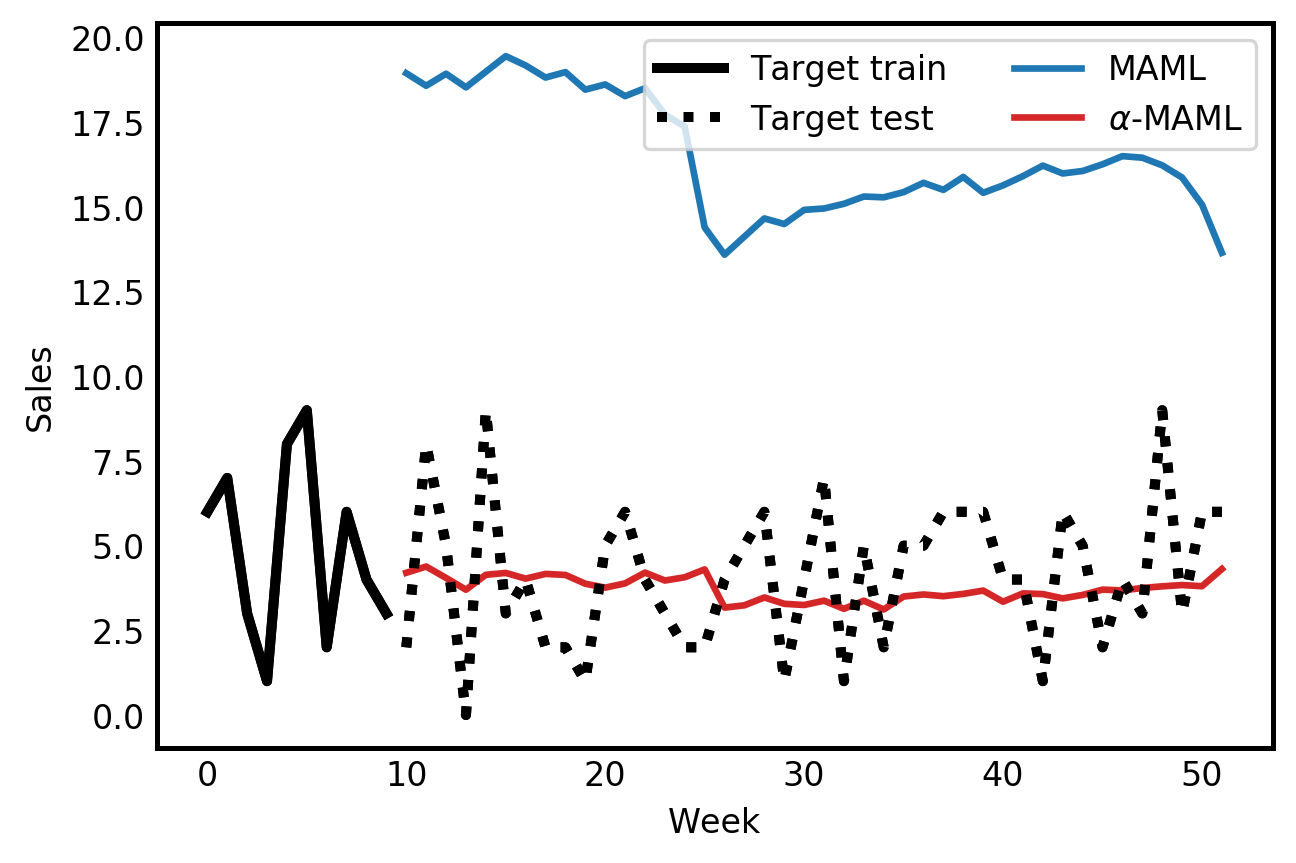}
    \caption{
        Product sales over 52 weeks.
        Example target task from the sales data set;
        the solid black line denotes the data examples
        used for training,
        and the dashed black line denotes the data used
        for testing.
        The blue and red lines denote the learned intializations (i.e.,
        before fast adaptation).
    }
    \label{fig-sales}
\end{figure}

In \Cref{fig-sales} we show the learned initializations from $\alpha$-MAML
vs uniformly-weighted MAML, where the learning rate parameter was set as $\eta = 0.0001$.
Here the MAML initialization learns a task that is an average of many of the tasks;
in constrast, the $\alpha$-MAML initialization
upweights products with more similar sources and patterns as the target.
As a result, the $\alpha$-MAML initization is able to better predict
future data coming from that task.

\section{Discussion and future work}

We presented a class of weighted meta-learning methods,
where the weights are selected by minimizing a data-dependent bound
involving an empirical IPM between the weighted sources and target risks.
Using this bound, we developed a computable algorithm based on minimizing an
empirical kernel distance, providing examples
for basis regression models with square loss and hinge loss.

A number of promising future directions remain.
One direction is to generalize our approach to
arbitrary loss functions, beyond the square and hinge loss,
and to extend the method to multi-class classification problems;
here it would be necessary to develop additional computational improvements.
Additionally, one could consider only use the labeled target task
examples during training, but also unlabeled target information to help quickly adapt the tasks.
Finally, exploring the use of this method in other applications, such as a continual learning
paradigm, remains a fruitful direction.

\subsubsection*{Acknowledgments}
This work was partially completed while Diana Cai was at Microsoft Research New England.
Diana Cai is supported in part by a Google Ph.D.\ Fellowship in Machine Learning.

\appendix

\section{Experimental details}
\label{appendi-experiments}

In this section, we present additional experimental
details and results to  complement the results presented in the main paper.
In \Cref{ssec-maml-soln}, we provide a derivation of
 the analytical solution of weighted MAML and ERM.
In \Cref{ssec-gen-bound}, we discuss an alternative
weighted meta-learning algorithm, giving by directly
optimizing a generalization bound, and explore the results
on the synthetic sine wave regression task.
Lastly, we present additional results and details
for the experiments considered in \Cref{sec-experiments}.

\subsection{The analytical $\alpha$-weighted meta-learning solution}
\label{ssec-maml-soln}

The solution to the weighted MAML (and weighted ERM)
meta-objective is available in closed form, as we show
in this section.
We assume a linear model and squared loss for every task.
We follow \citet[Appendix~A]{finn19a}, who provide a derivation
of the analytical solution for the uniformly-weighted case of ERM (i.e., joint
training) and MAML for linear regression with squared loss.

Denote the  MAML adaptation function of the predictors $w \in \reals^d$
as
\[U_j(w) := w - \eta (A_j w - b_j),\]
where $A_j := X_j^\top X_j$, $b_j := X_j^\top w$,
$X_j \in \reals^{N^{(j)}\times d}$ is the covariate matrix
of the $j$-th source task, and $\eta > 0$ is the step size.

The weighted MAML objective can be written as a function
of the predictors $w$ as follows
\begin{align*}
    F(w) &=
    \sum_{j=1}^J \alpha_j
    \left(\frac{1}{2} U_j(w)^\top A_j U_j(w) - U_j(w)^\top b_j\right)
    \\
    &=
    \frac{1}{2}
    w^\top\left(\sum_{j=1}^J \alpha_j (I - \eta A_j)^\top A_j (I - \eta
    A_j)\right) w
    \\
    &\quad +
    w^\top\left(\sum_{j=1}^J \alpha_j (I - \eta A_j)^\top b_j\right).
\end{align*}
Defining $\tilde A_j :=(I - \eta A_j)$, and
\begin{align*}
    \tilde A :=  \sum_{j=1}^J \alpha_j \tilde A_j^\top A_j \tilde A_j,
    \quad
    \tilde b := \sum_{j=1}^J \alpha_j \tilde A_j^\top b_j,
\end{align*}
we have that the gradient of the meta-objective is
\begin{align*}
    \nabla F(w) = \tilde A w - \tilde b,
\end{align*}
and so the solution is
    $w_{\alpha\text{-MAML}} = \tilde A^{-1} \tilde b.$

When the MAML learning rate $\eta=0$,
we recover the solution for the $\alpha$-weighted ERM,
where $U(w) = w$.

\subsection{Generalization bound optimization}
\label{ssec-gen-bound}

In \Cref{sec-algorithm} of the main paper, we describe a high-level
algorithm for optimizing for the $\alpha$ weight values, given by
\begin{align*}
    \hat \alpha := \argmin_{\alpha \in \Delta^{J-1}}
    \gamma_k(\sourcemixtureempdistrib, \targetempdistrib),
\end{align*}
i.e., minimizing a kernel distance between
the empirical distributions of the $\alpha$-mixture of sources and the target.

An alterative algorithm could also be derived from directly
optimizing a generalization bound.
Indeed, \Cref{thm-second} implies that
for all $g \in \G$,
\begin{align*}
    \E_\targetdistrib(g)
    \leq
    \E_{\sourcemixtureempdistrib}(g)
    +
    \ipm{\sourcemixtureempdistrib}{\targetempdistrib}{\fclass}
    +
    {2\mathcal{R}(\fclass|\point_1,\dots,\point_{\size{\targetindex}})}
    +
    3\sqrt{\frac{(b-a)^2\log(2/\epsilon)}{2\size{\targetindex}}},
\end{align*}
and
similarly, \Cref{corr} implies that for all $g \in \G$,
\begin{align*}
    \E_\targetdistrib(g)
    \leq
    \E_{\sourcemixtureempdistrib}(g)
    +
    \sum_{\sourceindex=1}^\numsources
    \alpha_\sourceindex
    \ipm{\sourceempdistrib}{\targetempdistrib}{\fclass}
    +
    {2\mathcal{R}(\fclass|\point_1,\dots,\point_{\size{\targetindex}})}
    +
    3\sqrt{\frac{(b-a)^2\log(2/\epsilon)}{2\size{\targetindex}}}.
\end{align*}

Thus, an alternative algorithm to the one
proposed in \Cref{algorithm-alpha} would involve
directly optimizing the generalization bound above:
\begin{align}
    \label{eq-direct-gen-bound}
    \hat \alpha, \hat{g} :=
    \argmin_{\alpha \in \Delta^{J-1}, g\in \G}
    \E_{\sourcemixtureempdistrib}(g) +
    \gamma_k(\sourcemixtureempdistrib, \targetempdistrib).
\end{align}

We can also propose a variant of the looser bound given by optimizing
\begin{align}
    \hat \alpha, \hat{g} :=
    \argmin_{\alpha \in \Delta^{J-1}, g \in \G}
    \E_{\sourcemixtureempdistrib}(g) +
    \sum_{\sourceindex=1}^\numsources
    \alpha_\sourceindex
    \gamma_k(\sourceempdistrib, \targetempdistrib).
\end{align}

An advantage of optimizing the generalization bound directly
rather than the two-step procedure in \Cref{algorithm-alpha}
is that we can jointly optimize for the $\alpha$ weight values
and model parameters via gradient descent.

\begin{figure*}[t!]
    \centering
    \includegraphics[scale=0.5]{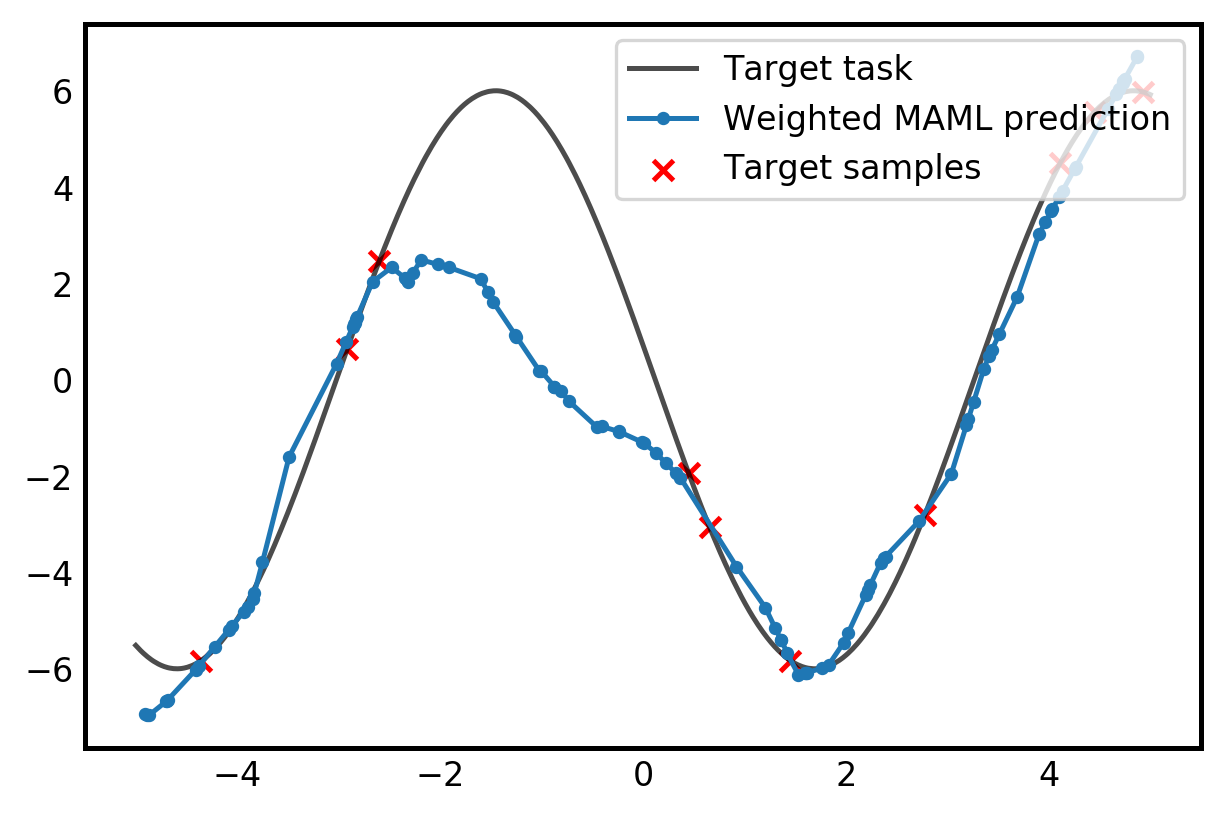}
    \includegraphics[scale=0.5]{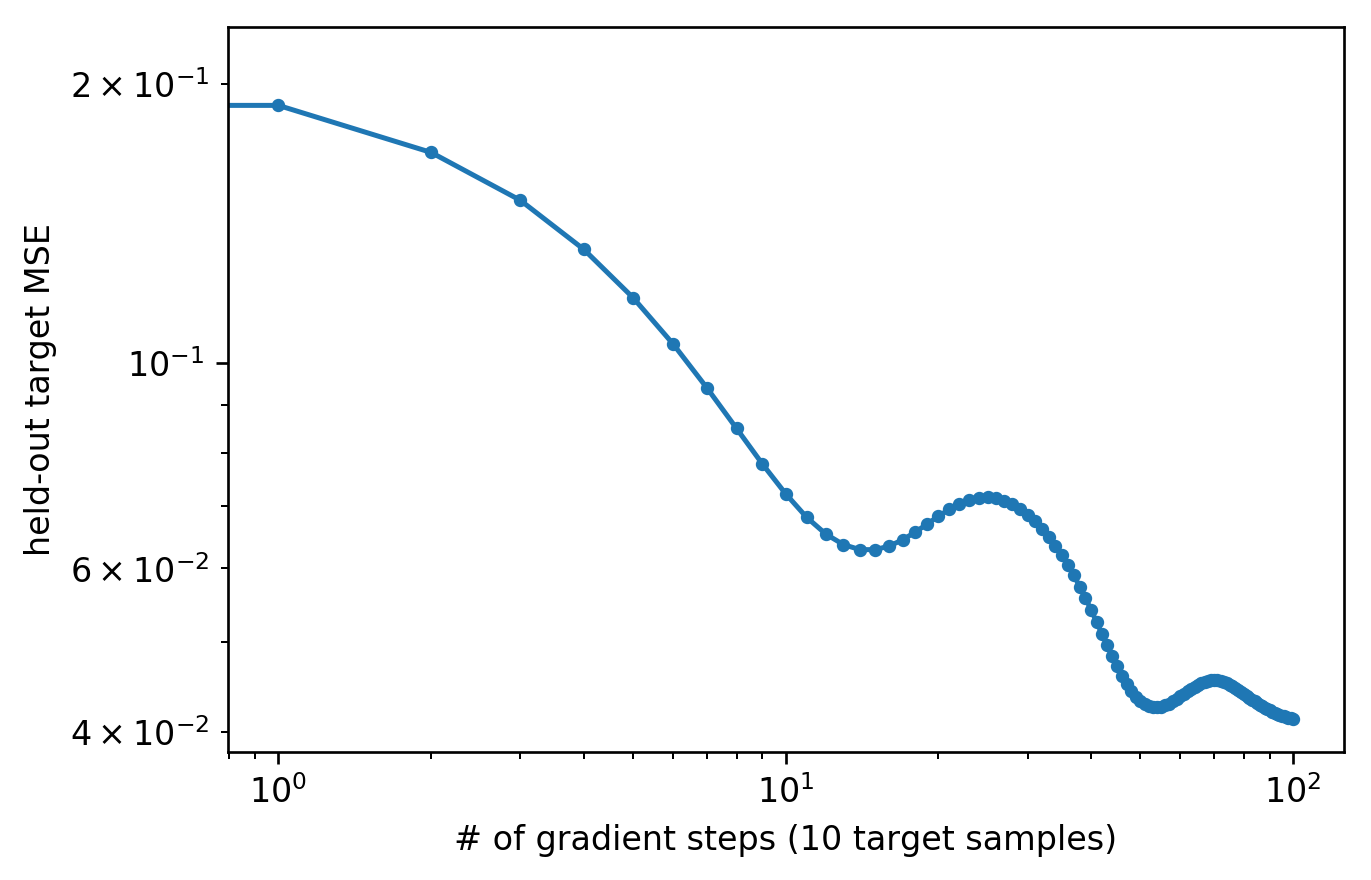}
    \caption
    {
        Results of sine wave regression
        obtained from directly optimizing the
        generalization bound in \Cref{eq-direct-gen-bound}
        during meta-training.
        \textbf{Left:}
        Learned weighted MAML initialization after 20,000 meta-iterations.
        \textbf{Right:}
        Held-out MSE ater fast adaptation using 10 samples of the target.
        }
    \label{fig-sine-direct-bound}
\end{figure*}

We examined the performance of the direct
bound optimization above, i.e.,
optimizing \Cref{eq-direct-gen-bound} in the
sine wave regression setting.
We sampled 200,000
source tasks in advance, according to the same task
distribution described in \Cref{sec-experiments},
and used the same 2-layer neural network model as before.
Adam was used for the meta-optimizer, with the same
learning rates of the main document.
In order to speed up the computation,
we used mini-batches of size 150.
Meta-training was performed for 20,000 meta-iterations.

In \Cref{fig-sine-direct-bound},
we show the results of the direct bound
optimization for the sine wave regression example.
The top plot shows that the initialization learned is
close to the target task,
though it does not seem to be able to capture areas
where there are no samples as well.
By contrast, while the predictions obtained
according to \Cref{algorithm-alpha}
(see main document, \Cref{fig-sine-regression})
are able to better capture the overall
shape of the sine wave task in only 10,000 meta-iterations.

The bottom plot shows that
the initialization is able to benefit from
fast adaptation, as it can be
adapted to the target with a small number of gradient steps;
however, the initialization obtained from \Cref{algorithm-alpha} is able to
adapt more quickly for this task.

Overall, this suggests that the convergence of the
direct bound optimization
is slower than the $\alpha$-MAML algorithm of
\Cref{algorithm-alpha} for the sine wave regression task.
However, the direct bound optimization
may still be of interest given enough computational
resources, as it is a simpler procedure to implement.

\subsection{Sine wave regression}

In the main paper, we explored sine regression
using an adaptive basis version of
\Cref{algorithm-alpha}.
In particular, we presented plots of a single
target and the resulting initializations
learned from MAML and $\alpha$-MAML.

Here we present additional results for
5-shot
target training sizes
using the adaptive version of \Cref{algorithm-alpha} (based on optimizing
an upper bound on \Cref{thm-second}.
We also evaluate a variant that we refer to as
the ``threshold'' meta-learning method,
that is based on optimizing an upper bound on
\Cref{corr},
which corresponds to weighting only
the closest source task.

In each random trial, a random sine task was drawn according to the task
distribution described in the main paper, and random samples from the target
were also drawn (with fixed amplitude and random phase).
Each method was trained for 10,000 meta-iterations.
The intializations are evaluated on 1000 held-out samples
from each sine task.

\begin{figure*}
    \centering
    \includegraphics[scale=0.5]{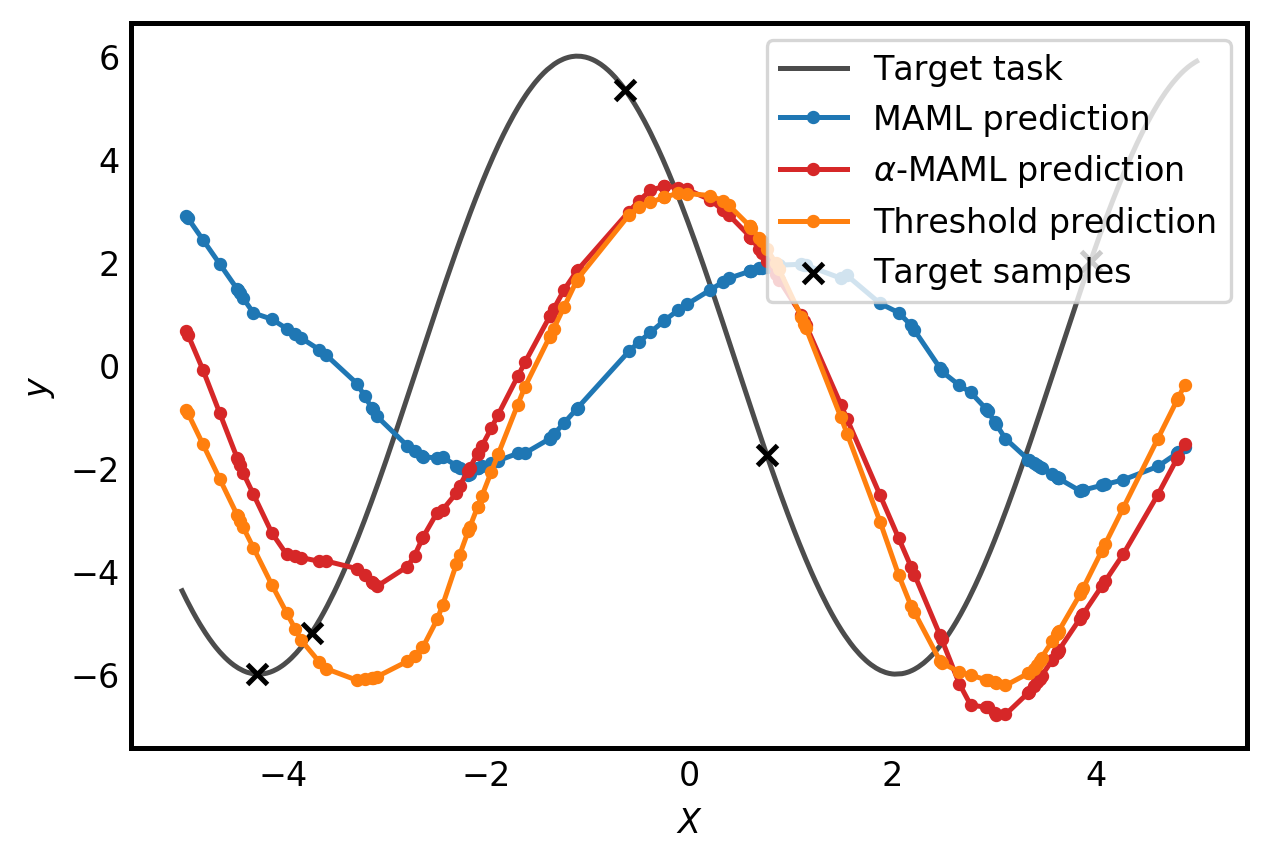}
    \includegraphics[scale=0.5]{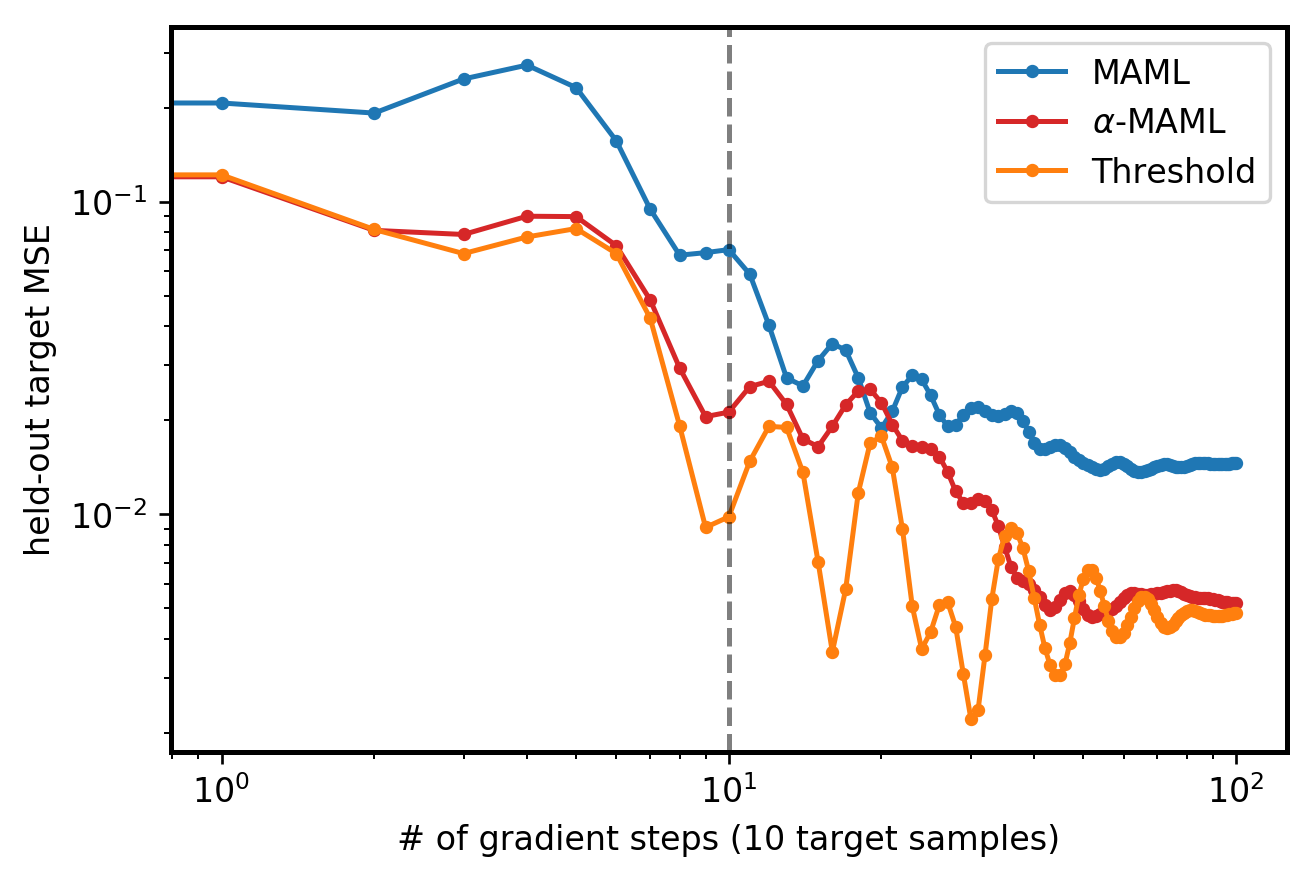}
    \caption
    {
        Results of sine wave regression for a 5-shot target task.
        \textbf{Left:}
        Learned meta-initializations for uniform, $\alpha$,
        and single-source weights after 10,000 meta-iterations.
        \textbf{Right:}
        Target task MSE on 100 held-out target samples after fast adaptation using 5 samples from the
        target.
    }
    \label{fig-sine-5shot}
\end{figure*}

In \Cref{fig-sine-5shot}, we present a single trial
from a 5-shot target task
with the learned initializations of each method
and the held-out MSE after fast adaptation.
In this example, for all methods,
fast adaptation helps, implying that the learned
meta-initialization is useful for learning
this task.
However,
even after a large number of target tasks,
the uniformly-weighted MAML
initialization is unable to achieve the same MSE as the non-uniformly
weighted initializations  (i.e., $\alpha$- and threshold-MAML).

\subsection{Multi-dimensional regression on real data sets}

In this section, we provide details on
how the data sets used for multi-dimensional
linear regression in the main document
were divided into separate source and target tasks.
We also visualize the inferred $\alpha$ weight values
that are used for the predictions reported in
\Cref{table-mult-regression},
where the $\alpha$-weighted methods provide
a small improvement in RMSE over the uniformly-weighted
methods.

\paragraph{Diabetes data.}

We split the diabetes data set into
multiple sources by grouping on the following age groups:
$[19,29)$,
$[29,39)$,
$[39,49)$,
$[53,59)$,
$[59,64)$,
$[64,79).$
The target age group included data from the
age group $[49,52]$.

\begin{figure*}[t]
    \centering
    \includegraphics[scale=0.5]{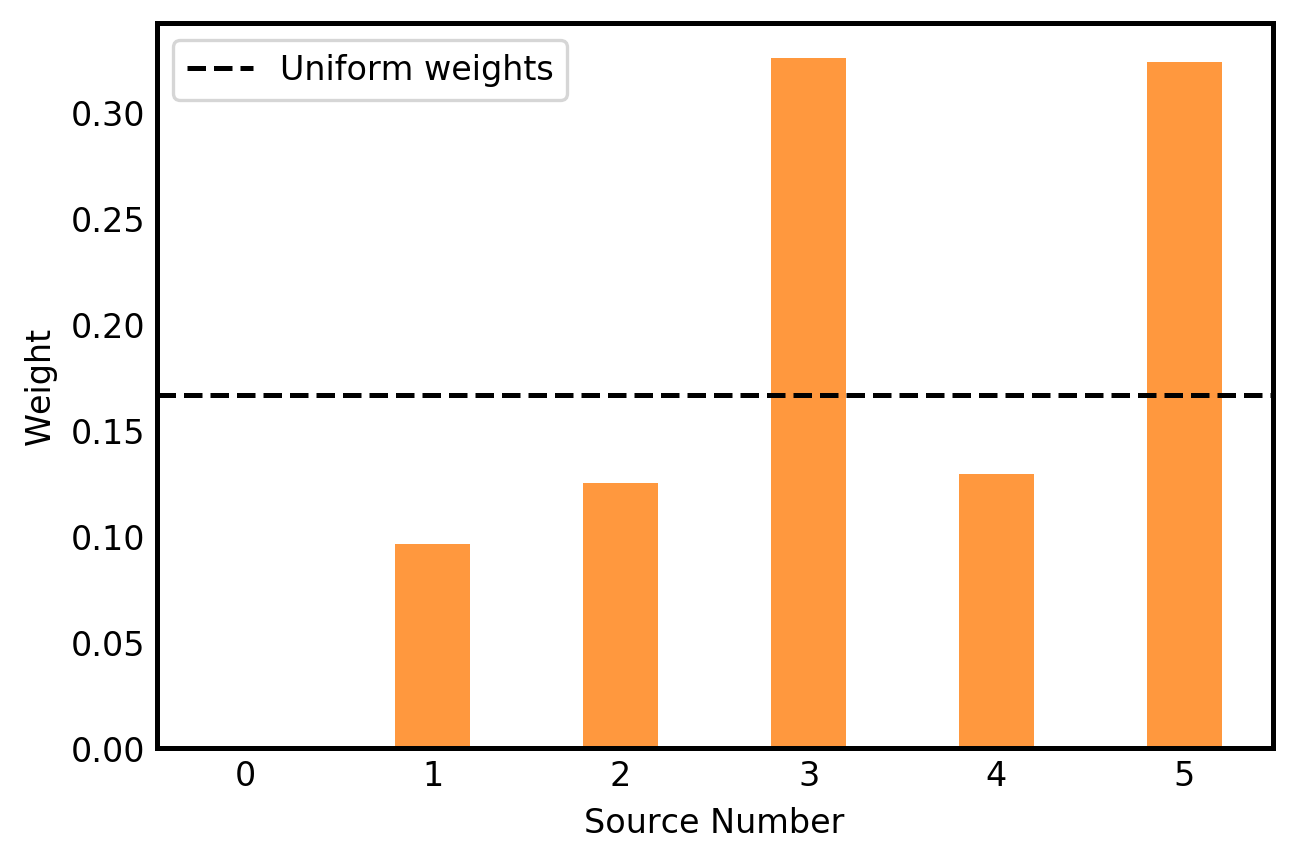}
    \includegraphics[scale=0.5]{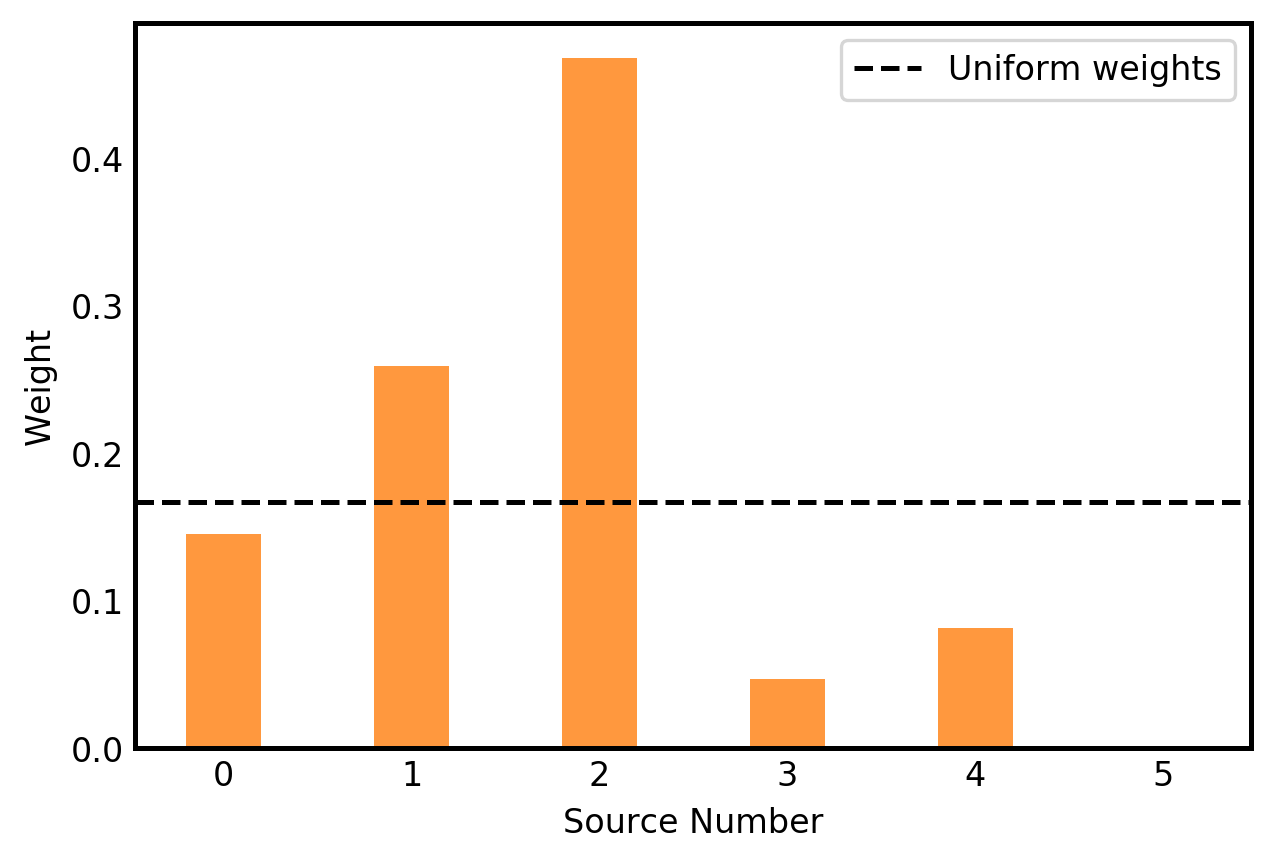}
    \caption{
        The learned $\alpha$-weighting according to
        the kernel IPM between the $\alpha$-mixture
        of sources and target.
        \textbf{Left}: Diabetes data set.
        \textbf{Right}: Boston data set.
        }
        \label{fig-diabetes}
\end{figure*}

\Cref{fig-diabetes} (left) shows a visualization of
the sources that were upweighted (blue) and
downweighted (gray).
In particular,
the sources from the age groups $[39,49)$
and $[64,79)$ were upweighted,
and the source corresponding to the age group of
$[19,29)$ received 0 weight,
which indicates that the  closer age groups, i.e.
$[39,49)$ and $[64,79)$ ,
are more similar sources to learn from
the younger age group $[19,29)$
for the target task (i.e., the age group $[49,52]$).

\paragraph{Boston housing prices.}

The Boston housing prices data set was split into
the sources by grouping the sources tasks
on the following age groups:
$[2.9, 29.1)$,
$[29.1, 42.3),$
$[42.3, 58.1),$
$[72.5, 84.4),$
$[84.4, 92.4),$
and
$[92.4, 100.0)$.
The target age group included data from the
age group
$[58.1, 72.5)$.

The inferred $\alpha$ weight values
are displayed in \Cref{fig-diabetes} (right).
Here we see that the sources corresponding
to the age groups
$[29.1, 42.3)$ and
$[42.3, 58.1)$ are upweighted,
while
the source corresponding to the age group
$[92.4, 100.0)$ received 0 weight.

\bibliographystyle{plainnat-mod}
\bibliography{main.bib}

\end{document}